\documentclass[journal]{IEEEtran}
\ifCLASSINFOpdf
\else
\fi
\usepackage{times}
\usepackage{soul}
\usepackage{url}
\usepackage[utf8]{inputenc}
\usepackage{caption}
\usepackage{amsmath}
\usepackage{booktabs}
\usepackage{colortbl}
\usepackage{xcolor}
\usepackage{graphicx}
\usepackage{booktabs}
\usepackage{multirow}
\usepackage{pifont}

\usepackage{longtable}
\usepackage{booktabs}
\usepackage{caption}
\usepackage{amssymb}  

\begin{document}

\title{Light4GS: Lightweight Compact 4D Gaussian Splatting Generation via Context Model}



\author{Mufan Liu, 
Qi Yang,~\IEEEmembership{Member,~IEEE,}
He Huang,
Wenjie Huang,
Zhenlong Yuan,
Zhu Li,~\IEEEmembership{Senior Member,~IEEE,}
Yiling Xu,~\IEEEmembership{Member,~IEEE,}
Yunfeng Guan
\thanks{
Mufan Liu, He Huang, Wenjie Huang, Yiling Xu and Yunfeng Guan are with the School of Information Science and Electronic Engineering, Shanghai Jiao Tong University, Shanghai 200240, China (e-mail: \{sudo\_evan, huanghe0429, huangwenjie2023, yl.xu, yfguan69\}@sjtu.edu.cn).

Qi Yang and Zhu Li are with the School of Science and Engineering, University of Missouri–Kansas City, Kansas City, MO 64110 USA (e-mail: \{qiyang, lizhu\}@umkc.edu).

Zhenlong Yuan is with the University of Chinese Academy of Sciences, Beijing 100049, China (e-mail: yuanzhenlong21b@ict.ac.cn).

Corresponding author: Qi Yang and Yiling Xu.}
}

\definecolor{darkgreen}{RGB}{0, 100, 0}    
\definecolor{darkpurple}{RGB}{128, 0, 128} 
\definecolor{darkblue}{RGB}{0, 0, 139}     


%

\maketitle

\begin{abstract}
3D Gaussian Splatting (3DGS) has emerged as an efficient and high-fidelity paradigm for novel view synthesis. To adapt 3DGS for dynamic content, deformable 3DGS incorporates temporally deformable primitives with learnable latent embeddings to capture complex motions. Despite its impressive performance, the high-dimensional embeddings and vast number of primitives lead to substantial storage requirements.  In this paper, we introduce a \textbf{Light}weight \textbf{4}D\textbf{GS} framework, called Light4GS, that employs significance pruning with a deep context model to provide a lightweight storage-efficient dynamic 3DGS representation. The proposed Light4GS is based on 4DGS, which is a typical representation of deformable 3DGS. Specifically, our framework is built upon two core components: (1) a spatio-temporal significance pruning strategy that eliminates over 64\% of the deformable primitives, followed by an entropy-constrained spherical harmonics compression applied to the remainder; and (2) a deep context model that integrates intra- and inter-prediction with hyperprior into a coarse-to-fine context structure, enabling efficient multiscale latent embedding compression. Our approach achieves over 12× compression and increases rendering FPS up to 20\% compared to the baseline 4DGS, and also superior to frame-wise state-of-the-art 3DGS compression methods. Experiment results show the effectiveness of our Light4GS in terms of both intra- and inter-prediction methods without sacrificing rendering quality. The code is available at https://github.com/Evan-sudo/Light4GS. 
\end{abstract}


\section{Introduction}
\label{sec:intro}

\begin{figure}[t]
    \centering
    \includegraphics[width=0.95\linewidth]{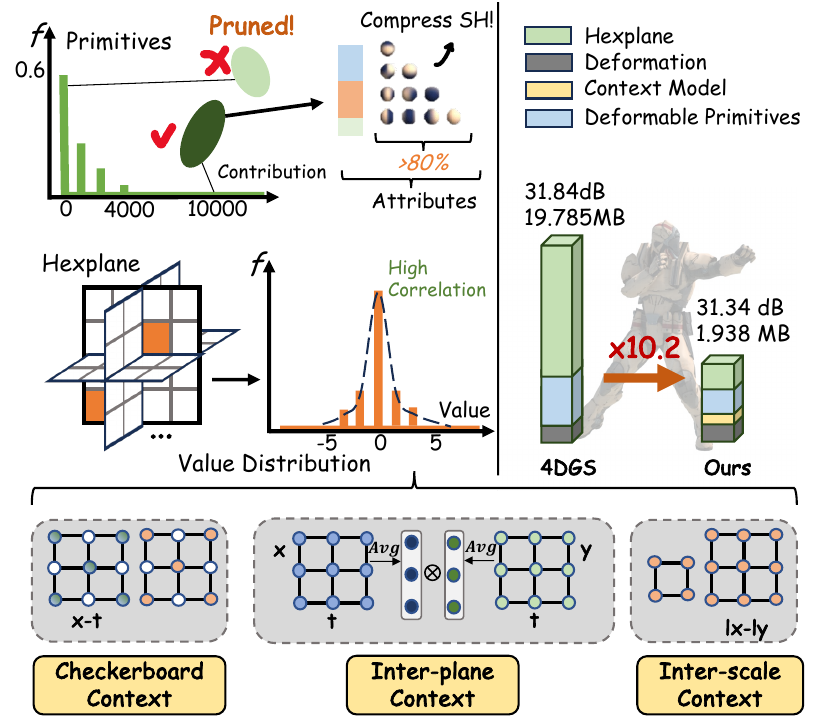}
    \caption{Motivation of 4DGS compression. Deformable primitives and hexplanes account for more than 99\% of 4DGS's storage (\textbf{top-right}). More than 60\% of primitives show almost 0 contribution to the final rendering and the hexplane exhibits strong inter-correlation (\textbf{top-left}). Considering these, we introduce STP to prune insignificant primitives and employ MHCM to compress hexplanes, with the three key technical components (\textbf{bottom}).}
    \label{preexp}
    \vspace{-0.5cm}
\end{figure}

 Dynamic view synthesis has become a central problem in 3D vision, with the goal of generating photorealistic renderings of dynamic scenes from arbitrary viewpoints \cite{Yuan2025a, yuan2025dvpmvssynergizedepthnormaledgeharmonized, Du2019Montage4D}. 3D Gaussian Splatting (3DGS) has recently gained significant attention for its ability to deliver faster rendering and higher visual quality than neural radiance field (NeRF)-based counterparts \cite{bao20253d} in terms of static view synthesis. It represents 3D scenes using anisotropic Gaussian primitives and is trained end-to-end via a highly efficient differentiable rasterization pipeline. Similar to NeRF, 3DGS can be extended with a temporal dimension to support dynamic view synthesis. 
 
 Recent efforts in dynamic 3DGS have led to two main categories of approaches: time-variant and deformation-based. Time-variant methods \cite{STG, ICLR} model Gaussian attributes as continuous functions of time, enabling shared representations across frames and avoiding per-frame training. Although impressive results are achieved, time-variant methods require substantial memory and incur considerable training overhead. Deformable 3DGS reduces modeling complexity by employing a lightweight multilayer perceptron (MLP) to deform primitives learned from a static 3DGS template. This idea is inspired by deformable NeRF approaches \cite{dnerf, park2021nerfies}, deformation of primitives is typically assisted by latent embeddings such as positional embeddings \cite{d3dgs, ed3dgs, huang2025adc}, hash tables \cite{xu2024grid4d4ddecomposedhash} and low-rank grids \cite{4DGS}. Among them, 4DGS \cite{4DGS}, which utilizes multiscale hexplanes as latent embeddings, is the most representative one. By mapping 4D space into six orthogonal planes, hexplane efficiently reduces computational complexity, making 4DGS more practical for real applications than the time-variant solution. Nonetheless, 4DGS’s storage requirements remain prohibitively high for real-world streaming. While current 5G networks stably support high-definition video at bitrates below 3 Mbps \cite{hevc, liu2024evan}, 4DGS requires 8–10 Mbps for comparable scenes. This gap shows that 4DGS is still too large for practical use in real-world streaming settings. 
 
 Fortunately, the two largest components of 4DGS show considerable redundancy and are prime candidates for compression. As illustrated in Fig. \ref{preexp}, over 60\% of its primitives have minimal impact on rendering and can be pruned without loss in quality. Moreover, multiscale hexplanes exhibit strong correlations, suggesting that deep context modeling could greatly mitigate their information redundancy. Motivated by these insights, we aim to achieve effective dynamic 3DGS compression based on the 4DGS anchor.

 In this paper, we propose \textbf{Light4GS}, a \textbf{light}weight \textbf{4}D \textbf{G}aussian \textbf{S}platting compression framework based on 4DGS \cite{4DGS}, which leverages efficient pruning and deep context modeling. The motivation behind our approach is illustrated in Fig. \ref{preexp}. To improve spatio-temporal encoding, we first apply Principal Component Analysis (PCA) to the spatial coordinates of the Gaussian primitives, aligning them with the principal directions of the scene. Then, a non-linear contraction is applied to the spatial coordinates to address the unbounded nature of 4DGS, ensuring that queried points remain within the hexplane boundaries. Next, our framework compresses both the canonical 3DGS and its associated hexplane features: 1) For canonical 3DGS, we introduce a Spatio-Temporal Significance Pruning (STP) strategy to globally remove unimportant primitives. STP evaluates each primitive’s contribution across multiple views and timestamps, and prunes deformable primitives based on their significance ranking. This also accelerates training by skipping gradient updates for pruned primitives and their associated hexplane features. Since spherical harmonics (SH) coefficients account for over 80\% of the primitive's data volume, entropy-constrained compression is further applied to compress the SH. 2) For hexplane compression, we propose a Multiscale Hexplane Context Model (MHCM) by leveraging both intra-scale and inter-scale redundancies in a coarse-to-fine manner. At the lowest scale, we adopt a checkerboard context model combined with either a learned hyperprior or an inter-plane context, depending on the plane type. Space-time planes are encoded first using latent hyperpriors extracted from a deep encoder. The decoded results are then time-averaged to form an inter-plane context, which guides the compression of space-only planes without introducing extra side information. At higher scales, we introduce inter-scale context by decoding finer hexplanes using bilinearly interpolated features from the preceding coarser scale. This design captures cross-scale redundancy, improves compression efficiency, and reduces training complexity. In summary, our contributions are as follows:
\begin{itemize}
    \item We propose Light4GS, a lightweight 4DGS compression framework that integrates primitive pruning with a deep context model for storage-efficient dynamic view synthesis.
    
    \item We introduce a spatio-temporal significance pruning strategy to remove redundant primitives, and apply entropy-constrained optimization to compress the spherical harmonics (SH) coefficients of the remaining ones. We design a multiscale Hexplane Context Model (MHCM) that efficiently compresses hexplane features by leveraging both intra- and inter-scale redundancies in a coarse-to-fine manner with adaptive quantization.
    
    \item Experiments on both synthetic and real-world datasets demonstrate that Light4GS achieves 10--200$\times$ compression over state-of-the-art (SOTA) dynamic neural view synthesis methods without compromising rendering quality.
\end{itemize}



\section{Related Works}
 \subsection{Dynamic view synthesis }Dynamic view synthesis aims to generate novel views of dynamic scenes from 2D images captured across different time instances, which remains a challenging problem \cite{Yuan2025b, Hu2025Thing2Reality}. Early approaches extend NeRF \cite{nerf} to 4D by either deforming a canonical neural radiance field or explicitly modeling it as time-varying \cite{dnerf, park2021nerfies, nerfplayer, ffdnerf}. While effective, these methods suffer from slow rendering speeds due to their computationally expensive per-ray queries. To overcome this, recent work introduces dynamic 3DGS, which extends 3DGS into the temporal domain. These methods generally fall into two categories: time-variant and deformation-based. Time-variant methods encode temporal changes directly into Gaussian attributes using polynomial, Fourier, or learned functions \cite{ dynamic3DGS, STG}, or by explicitly extending 3D Gaussian primitives into 4D space \cite{ICLR}. Although these methods often achieve high reconstruction quality, they incur significant training and storage costs. In contrast, deformation-based methods \cite{4DGS, dn4DGS, ed3dgs} represent a scene using a canonical set of 3D Gaussians and a deformation field to capture temporal variations. While such deformation fields may struggle to perfectly track highly dynamic objects, they are more compact and easier to train. These deformation models are often enhanced with neural feature encodings, such as hash grids \cite{HAC, muller2022instant}, per-Gaussian embeddings \cite{ed3dgs}, or low-rank grids \cite{TensorRF, HexPlane}. A representative example is 4DGS \cite{4DGS}, which combines a canonical Gaussian set with a Hexplane-based deformation field, achieving competitive performance with relatively low storage on both multi-view and monocular video datasets. We adopt it as the anchor design for our 4DGS compression method.

\subsection{3DGS and its Compression} 3DGS has emerged as a powerful alternative for 3D scene representation endowed with learnable shape and anisotropic attributes. The implementation is efficient as it only requires hundreds of multiview photos for a complex scene. By adopting differentiable splatting and tile-based rasterization, 3D Gaussians are optimized during training to best fit their local 3D regions. Its explicit structure also leads to faster training and rendering speeds than SOTA NeRF algorithms. Recent studies in 3DGS compression can be categorized into various categories. Post-processing methods, generally applied after training, reduce Gaussian primitives and compress their attributes utilizing neural network-inspired compression. Key strategies include significance pruning \cite{lightGS, compactGS, liu2025d2gv}, codebooks \cite{compactGS}, knowledge distillation \cite{lightGS}, and quantization \cite{niedermayr, yang2025hybridgs}. 
Thanks to the explicit structure of 3DGS, classical point cloud compression techniques such as octree encoding \cite{octreeGS, huang2024hierarchicalcompressiontechnique3d}, graph signal processing \cite{graph}, and region-adaptive hierarchical transform (RAHT) \cite{MesonGS} have been adapted for 3DGS compression with structural priors. A different line of research focuses on leveraging spatial relations to impose structured organization on otherwise unstructured Gaussians. These methods often collapse dimensions to arrange Gaussian primitives into regular 2D grids, facilitating efficient encoding. For example, SUNDAE \cite{yang2024spectrally} combines pruning strategies with spectral graph analysis to guide the creation of compact models. Mini-Splatting \cite{fang2024mini} proposes a spatially aware spawning mechanism to regulate the expansion of primitives. In contrast, Scaffold-GS \cite{scaffold} uses an anchor-based architecture, where Gaussian attributes are predicted via neural inference, reducing the number of parameters while improving reconstruction quality. Meanwhile, learning-based methods utilize VAE-style image compression frameworks that integrate hyperpriors and context modeling \cite{HAC, compgs} for more effective 3DGS compression.
\\

\subsection{Latent Embedding Compression} NVS methods utilizing latent embeddings \cite{4DGS, BiNeRF, muller2022instant, kplanes, HexPlane} achieve high-quality results, but also lead to increased storage overhead. To make embeddings smaller, \cite{muller2022instant} maps a large feature grid into a compact hash table, while \cite{BiNeRF} distills a high-dimensional feature grid into a binary grid. In dynamic NVS, storage overhead for latent embeddings becomes even more pronounced; low-rank approximations, such as wavelet decomposition \cite{wavelet}, vector-matrix decomposition \cite{TensorRF}, and HexPlane projection \cite{HexPlane, kplanes}, provide promising solutions by mitigating dimensional redundancy. Moreover, contextual dependencies among neighboring elements provide further compression potential, as explored in image compression  \cite{checkerboard, elic}, as well as in video compression \cite{dcvc, ma2019image}. To extend these advances into 3D representations, \cite{CNC, HAC} introduce spatial and hash-grid contexts to capture spatial redundancy in NeRF and 3DGS latent embeddings, which achieve nearly 100 compression gain over their vanilla implementations.\\
Building on 3DGS and latent embeddings compression, our approach benchmarks 4DGS compression by training lightweight deformable primitives with a deep context model to fully exploit spatio-temporal redundancy.

\section{Preliminary}
\textbf{3DGS: }3DGS represents a 3D scene using a collection of anisotropic Gaussian primitives distributed in space. Each primitive is characterized by a mean position $\boldsymbol{\mu} \in \mathbb{R}^3$ and a covariance matrix $\Sigma \in \mathbb{R}^{3 \times 3}$, defined as:

\begin{equation}
\mathbf{G}(\boldsymbol{x}) = e^{-\frac{1}{2} (\boldsymbol{x} - \boldsymbol{\mu})^\top \Sigma^{-1} (\boldsymbol{x} - \boldsymbol{\mu})}.
\end{equation}
For differentiable rasterization, the covariance matrix $\Sigma$ is decomposed into a rotation matrix $\mathbf{R}$ and a scaling matrix $\mathbf{S}$.
Each primitive is characterized by its color $c_i \in \mathbb{R}^k$ (parameterized by $k$ SH coefficients) and opacity $\alpha \in [0, 1]$. During rendering, the primitives are splatted onto the viewing plane, and the final pixel color $C$ is computed by blending the splatted primitives as:
\begin{equation}
C = \sum_{i=1}^{N}  c_i \alpha_i \prod_{j=1}^{i-1} (1 - \alpha_j),
\end{equation}
where $N$ is the number of primitives contributing to that pixel.

\begin{figure*}[t]
    \centering
    \includegraphics[width=1.0\linewidth]{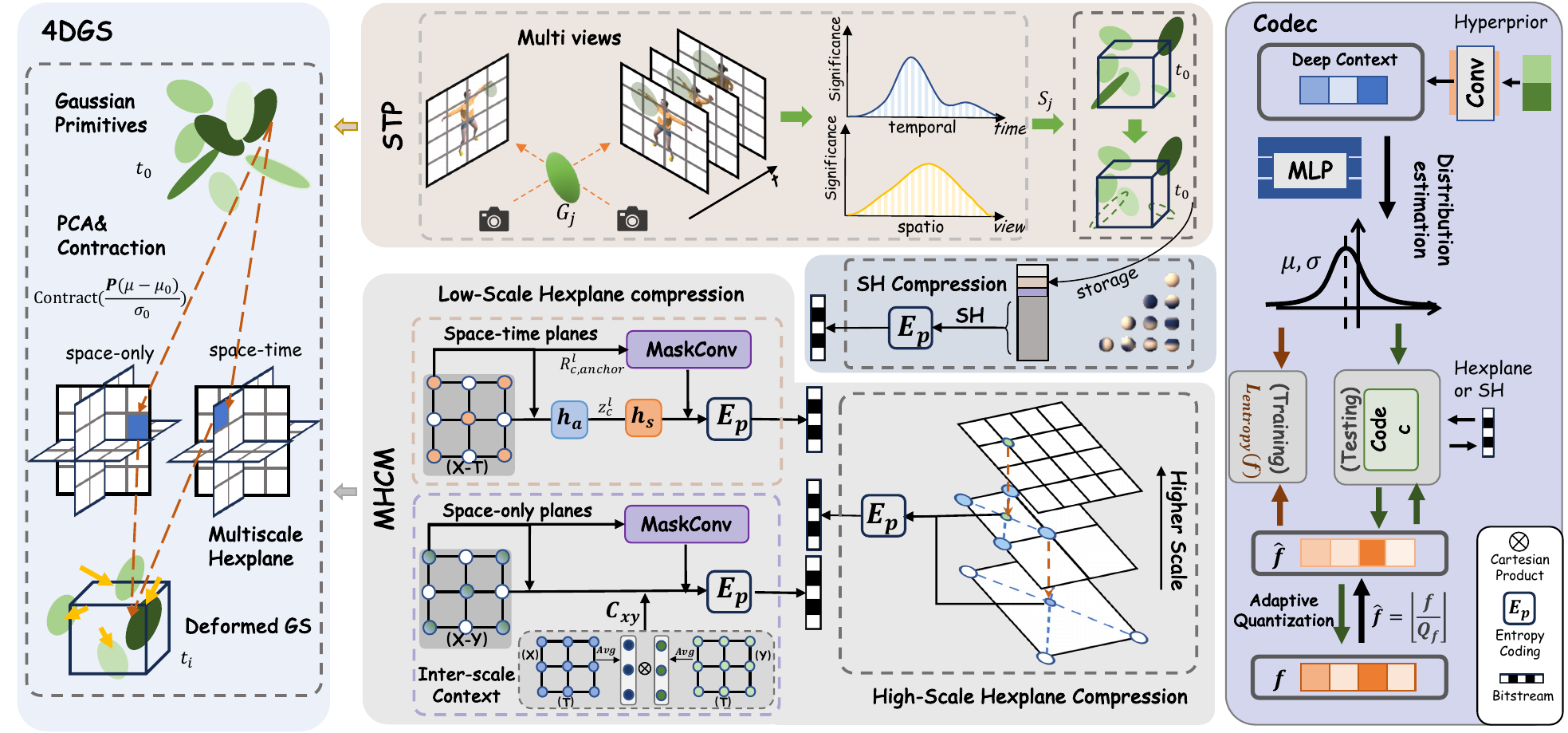}
    \caption{\textbf{Structure of Light4GS.} \textbf{Left:} Illustration of 4DGS. \textbf{Middle: }Storage reduction is achieved through \textbf{STP} and \textbf{MHCM}. Deformable primitives are pruned based on their spatio-temporal significance, while MHCM compresses multiscale hexplanes: the lowest-scale uses a checkerboard model with hyperprior or inter-plane context, and high-scales use inter-scale context. SH coefficients of remained primitives are further compressed via entropy-constrained coding. \textbf{Right:} Entropy-constrained codec pipeline, where deep context or hyperprior infers distribution prediction to encode/decode items $\boldsymbol{f}$ (multiscale hexplanes or SH). SH coding estimates distribution without any context.}
    \label{overview}
    \vspace{-0.5cm}
\end{figure*}

\noindent\textbf{4DGS:} To handle dynamic scenes with temporal variations, deformable 3DGS, i.e. 4DGS, extends 3DGS by deforming its static primitives. The key idea is to compute the canonical-to-world mapping of each primitive over time by a deformation network $\Phi$:
\begin{equation}
\Delta \mathbf{G} = \Phi(f_h, t).
\end{equation}
Here, $f_h$ represents the hexplane feature of primitive $\mathbf{G}$. Each primitive deformed at $t$ is computed by adding its deformation as:
\begin{equation}
\mathbf{G}' = \mathbf{G} + \Delta \mathbf{G}.
\end{equation}
As mentioned earlier, learnable multiscale hexplanes are introduced in 4DGS to better capture dynamic variations and assist deformation prediction (see Fig.\ref{overview}). Specifically, it factorizes four dimensions in a deformation field into six planes at different scales: $\mathbf{R}^l_c \in \mathbb{R}^{h \times lN_1 \times lN_2},\quad c \in \binom{{x, y, z, t}}{2}$, where $h$ is the hidden dimension of the feature, $(N_1, N_2)$ is the basic resolution, $l$ is the scaling factor and $c$ is the pair-wise dimension combination. 
Each primitive’s coordinates $\boldsymbol{\mu}$ are combined with timestamp $t$ to query each feature plane $\mathbf{R}^l_c$, where the four nearest features are retrieved and bilinearly interpolated. The interpolated features from all planes are element-wise multiplied, and concatenated from multiple scales to form the final hexplane feature $f_h$:
\begin{equation} f_h = \bigcup_{l} \prod \text{interp}\left( \mathbf{R}^l_c, (\boldsymbol{\mu}, t) \right), \quad c \in \binom{{x, y, z, t}}{2}.
\label{query}
\end{equation}
‘interp’ denotes the bilinear interpolation for querying the coordinate $(\boldsymbol{\mu}, t)$ located at 4 vertices of the feature plane.


\section{Method}
We present Light4GS, a storage-efficient deformable 3DGS compression framework that combines STP to reduce deformable primitives with MHCM for compressing multiscale hexplanes, as shown in Fig. \ref{overview}. We first present modifications to the vanilla 4DGS representation in Sec. \ref{modquery}. Then, we introduce our pruning strategy, i.e., STP, in Sec. \ref{STP}, followed by a detailed description of our deep context model and its coding process, i.e., MHCM, in Sec. \ref{context}. The remaining sections address our optimization techniques and loss functions.

\subsection{PCA-guided Hexplane Query for Unbounded 3DGS}
\label{modquery}
\begin{figure}[t]
    \centering
    \includegraphics[width=\linewidth]{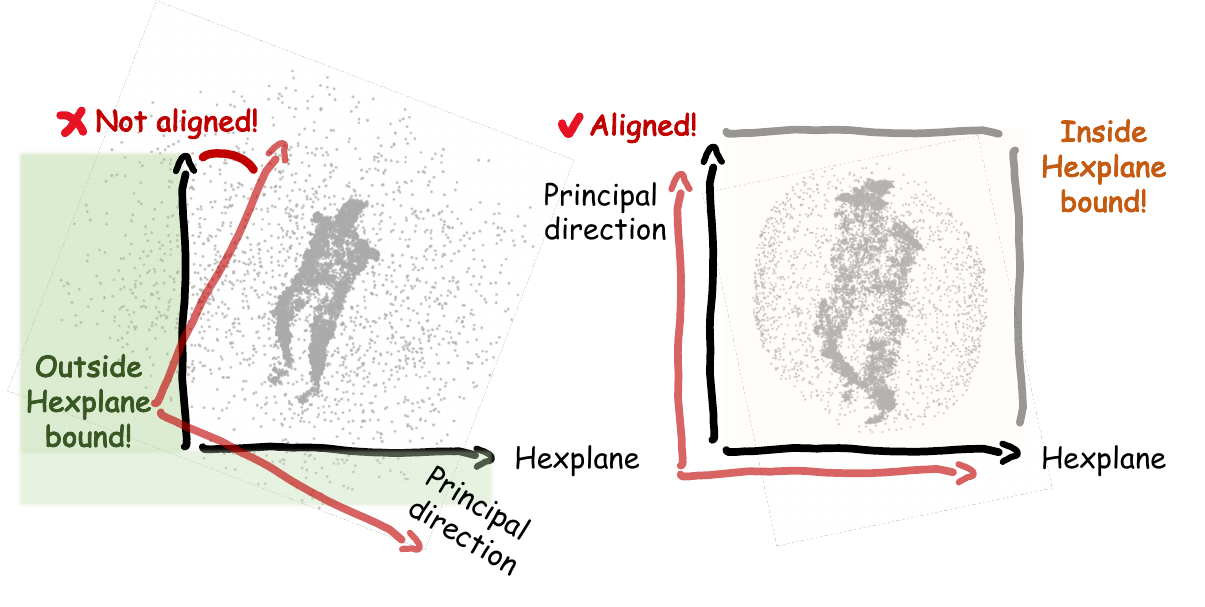}
    \caption{Gaussian splatting is unbounded, leading to queries outside the hexplane. Misaligned hexplane query directions with principal directions further reduce the amount of information the hexplane can learn (left); Nonlinear contraction constrains all points within the hexplane bounds, while PCA aligns query directions with principal directions (right).}
    \label{pca}
    \vspace{-0.3cm}
\end{figure}

The vanilla 4DGS encodes spatio-temporal variations using six orthogonal planes spanning all pairwise combinations of $(x, y, z, t)$. Despite this compact representation, we observe two key limitations (see Fig.~\ref{pca}): (1) while Gaussian primitives may occupy an unbounded 3D space, the hexplane grid is typically initialized within a fixed bounding box derived from the SfM point cloud, leading to a mismatch and uneven coverage of spatio-temporal features; and (2) the canonical axes $(x, y, z)$ are not necessarily aligned with the true principal directions of the scene, so the resulting non-principal projections can hinder the hexplanes from learning more informative latent features.

Motivated by how NeRF handles its unbounded issues, we adopt the non-linear spatial contraction of \cite{barron2022mip}:
\begin{equation}
\text{contract}(\boldsymbol{\mu}) = 
\begin{cases}
\boldsymbol{\mu}, & \|\boldsymbol{\mu}\| \leq 1 \\
\left(2 - \frac{1}{\|\boldsymbol{\mu}\|} \right)\frac{\boldsymbol{\mu}}{\|\boldsymbol{\mu}\|}, & \|\boldsymbol{\mu}\| > 1,
\end{cases}
\end{equation}
to map Gaussian centers $\boldsymbol{\mu}$ into a bounded spatial domain. Since the temporal dimension $t$ is already finite, no contraction is applied to it.

For the second issue, we perform PCA on the spatial coordinates $\boldsymbol{\mu}$ to align them with the principal directions, before applying the nonlinear contraction:
\begin{equation}
\boldsymbol{\mu}' = \text{contract} \left( \frac{(\mathbf{P} (\boldsymbol{\mu} - \boldsymbol{\mu}_0)}{\boldsymbol{\sigma}_0} \right),
\end{equation}
where $\mathbf{P} \in \mathbb{R}^{3 \times 3}$ is the PCA rotation matrix, and $\boldsymbol{\mu}_0, \boldsymbol{\sigma}_0$ and denote the mean and the variance vectors along the principal axes. 

Using the transformed coordinates $\boldsymbol{\mu}'$, we perform bilinear interpolation over the learnable multiscale hexplanes $\mathbf{R}^l_c \in \mathbb{R}^{h \times lN_1 \times lN_2}$, where $c \in \binom{\{1,2,3,4\}}{2}$ are PCA axis indices. The final hexplane feature $f_h$ is obtained by aggregating interpolated features across all planes and scales:
\begin{equation}
f_h = \bigcup_{l} \prod_{c} \text{interp}\left( \mathbf{R}^l_c, (\boldsymbol{\mu}',t) \right),
\end{equation}
This replaces the original query procedure in Eq. (\ref{query}).

\subsection{STP: Spatio-temporal Significance Pruning}
\label{STP}
Vanilla 4DGS reduces computation and storage overhead by pruning low-opacity primitives during training. However, this approach overlooks each primitive's \textit{impact range} on view rendering; high-opacity primitives, for instance, might affect only a few pixels. Additionally, primitives that cover a large portion of pixels in one frame may contribute minimally in the next. Ignorance of these factors leads to limited effectiveness of the primitive pruning.

 For this concern, a critical challenge in 4DGS is accurately capturing the global impact of each primitive on rendering quality. In 4DGS, each primitive is capable of deforming over time to account for scene dynamics, and this deformability enables us to assess the significance of one primitive across multiple frames. Building on this capability, we propose STP that evaluates each Gaussian's contribution by calculating its intersection with viewing pixels over all timestamps. The global significance score of the $j$-th deformable primitive $\text{S}_j$ is defined by aggregating its contributions across all timestamps and views as:
\begin{equation}
\text{S}_j = \sum_{t=1}^{\text{T}}\sum_{i=1}^{\text{MHW}} 
\mathbf{1}\left(\mathbf{G}+\Phi(f_h, t)),\ r_{i,t}\right) 
\cdot T_{j,i,t}
\cdot \sigma_j \cdot \gamma(\Sigma_{j,t}),
\end{equation}
where $\text{M}$ is number of views at each timestamp, $\text{H}, \text{W}$ are height and width of the rendered view. $\mathbf{1}(\cdot)$ is an indicator function that returns $1$ if the deformed primitive intersects with the viewing ray $r_{i,t}$ at timestamp $t$, and $0$ otherwise. {$T_{j,i,t}=\prod_{k<j}(1-\alpha_{k,i,t})$ denotes the per-pixel transmittance in alpha compositing, where $\alpha_{k,i,t}$ is the per-pixel opacity contributed by primitive $k$ at pixel $(i,t)$. $\sigma_j$ is the opacity of the primitive, and $\gamma(\Sigma_{j,t})$ is the normalized volume of primitive $j$  calculated from its covariance matrix $\Sigma_{j,t}$ at time $t$. During training, we periodically rank deformable primitives according to their spatio-temporal significance scores, $\text{S}_j$, and prune the least significant ones based on a predefined ratio. This approach greatly improves training efficiency by reducing the number of primitives to be processed, while optimizing the remaining primitives. Consequently, the retained primitives adapt to mitigate any potential quality loss from STP, preserving rendering quality despite the reduced model complexity.

\subsection{MHCM: Multiscale Hexplane compression with Context Model}
\label{context}
We use a deep context model to reduce information uncertainty within multiscale hexplanes to achieve a minimized entropy loss. Specifically, each hexplane feature is modeled as a Gaussian distribution conditioned on a carefully designed deep context. This context-conditioned distribution is parameterized by a deep network (denoted as MLP() here for simplicity), which takes the context as input to predict each feature’s distribution parameters. These parameters are then used to compute the entropy of hexplane features, and subsequently encode features using Arithmetic Coding (AC) \cite{langdon1984arithmetic}. The total entropy is incorporated into the 4DGS optimization to balance the rate-distortion (RD) trade-off. In Light4GS, we propose MHCM, which leverages intra- and inter-scale redundancies to compress multiscale hexplanes. Due to the mirroring nature of encoding and decoding, we present the context model from the encoding perspective, elaborating it in a coarse-to-fine manner.

\subsubsection{Lowest-Scale Hexplane Compression}Planes in one hexplane can be divided into two groups based on their dimensions: \textit{space-only} and \textit{space-time}. Throughout the lowest-scale encoding process, both groups employ checkerboard modeling but use different hyperpriors based on the availability of prior information. In this approach, we designate odd positions on each plane as anchors, which are encoded with only the hyperprior. In contrast, non-anchors (even positions) are encoded using both the hyperprior and the checkerboard context that is derived by applying masked convolution on the plane anchors.
Without loss of generality, we omit the superscript \( l \) (i.e., scaling factor in Eq. \eqref{query}) in this section, as it is always equal to 1. To compress the space-time group, since no prior information is available, we directly use a deep network \( h_a() \) to extract hyperpriors as \( \mathbf{z}_c = h_a(\mathbf{R}_{c}) \). The hyperprior is then applied to encode space-time planes as:

\begin{equation}
\mathbf{\Phi}_{c} = 
\begin{cases} 
\text{MLP}(h_s(\mathbf{z}_c), \textbf{0}), & \text{if }\mathbf{R}_{c} \in \text{anchors,}\\
\text{MLP}(h_s(\mathbf{z}_c), \text{MaskConv}(\mathbf{R}_{c})), & \text{otherwise.} 
\end{cases}
\label{checkerboard}
\end{equation}
$\mathbf{\Phi}_{c}$ is the distribution parameters for $\mathbf{R}_{c}$, $h_s()$ is the hyperprior decoding network and `MaskConv' denotes applying convolution on the anchors. We \textbf{prioritize} encoding the space-time planes, as they capture both spatial and temporal information. Once decoded, their time-averaged values provide the \textit{inter-plane context}, which serves as the hyperprior for the space-only planes. This eliminates the need for additional hyperprior storage and aids with decoded spatial information. To obtain the inter-plane context, consider an example where \( c = xy \) as shown in Fig. \ref{overview}; each inter-plane context at \( (x, y) \) is obtained by first computing the time-average of features at \( (x, t) \) and \( (y, t) \), then taking their Cartesian product.
The inter-plane context \(\mathbf{C}_c\) is constructed to match the dimensionality of the decoded hyperprior, \(\text{MLP}(h_s(\mathbf{z}_c))\), and thus replaces the decoded hyperprior in Eq. (\ref{checkerboard}) for coding space-only planes.


\subsubsection{High-Scale Hexplane Compression}
High-scale hexplanes often exhibit significant redundancy with their lower-scale counterparts. Relying solely on the checkerboard model for compression overlooks these inter-scale redundancies, as it captures only intra-scale correlations. To address this issue, we introduce an \textit{inter-scale context} that encodes higher-scale planes in a coarse-to-fine order. Specifically, higher-scale planes are decoded using context obtained from bilinear interpolation of its predecessor, as illustrated in Fig. \ref{overview}. The distribution parameters for higher-scale features \(\mathbf{R}^{l+1}_{c}\) are predicted as:
\begin{equation}
\mathbf{\Phi}_{c}^{l+1} = \text{MLP}(\text{interp}(\mathbf{R}^l_c, \mathbf{R}^{l+1}_c), \mathbf{0}).
\end{equation}
Here, 'interp' denotes bilinear interpolation to query the coordinates \(\mathbf{R}^{l+1}_c\) located at the four vertices of \(\mathbf{R}^{l}_c\). This progressive refinement greatly reduces redundancy and simplifies the deep context training. 
\begin{figure}[t]
    \centering
    \begin{minipage}{0.48\linewidth}
        \centering
        \includegraphics[width=\linewidth]{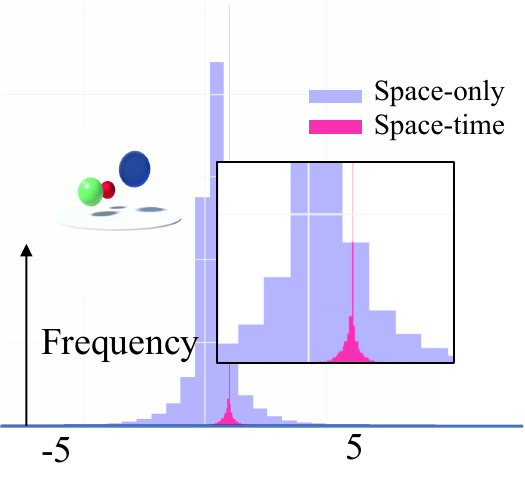}
        \label{fig:enter-label}
    \end{minipage}%
\begin{minipage}{0.5\linewidth}
    \centering
    \scriptsize  
    \begin{tabular}{c c}
      \toprule[1pt]
        4DGS & 40.12 dB \\
     \midrule
       \cellcolor{red!30}{w/ AQ} &    \cellcolor{red!30}{39.94} dB \\
        w/o AQ & 39.27 dB \\
     \bottomrule[1pt]
    \end{tabular}
    \caption*{\scriptsize PSNR Comparison on \textit{Bouncingballs}.}
\end{minipage}
\vspace{-0.8cm}
\caption{Distribution of space-time and space-only planes. Applying AQ achieves rendering quality improvement.}
\label{prune}
\vspace{-0.4cm}
\end{figure}

\begin{figure*}
    \centering
    \includegraphics[width=0.99\linewidth]{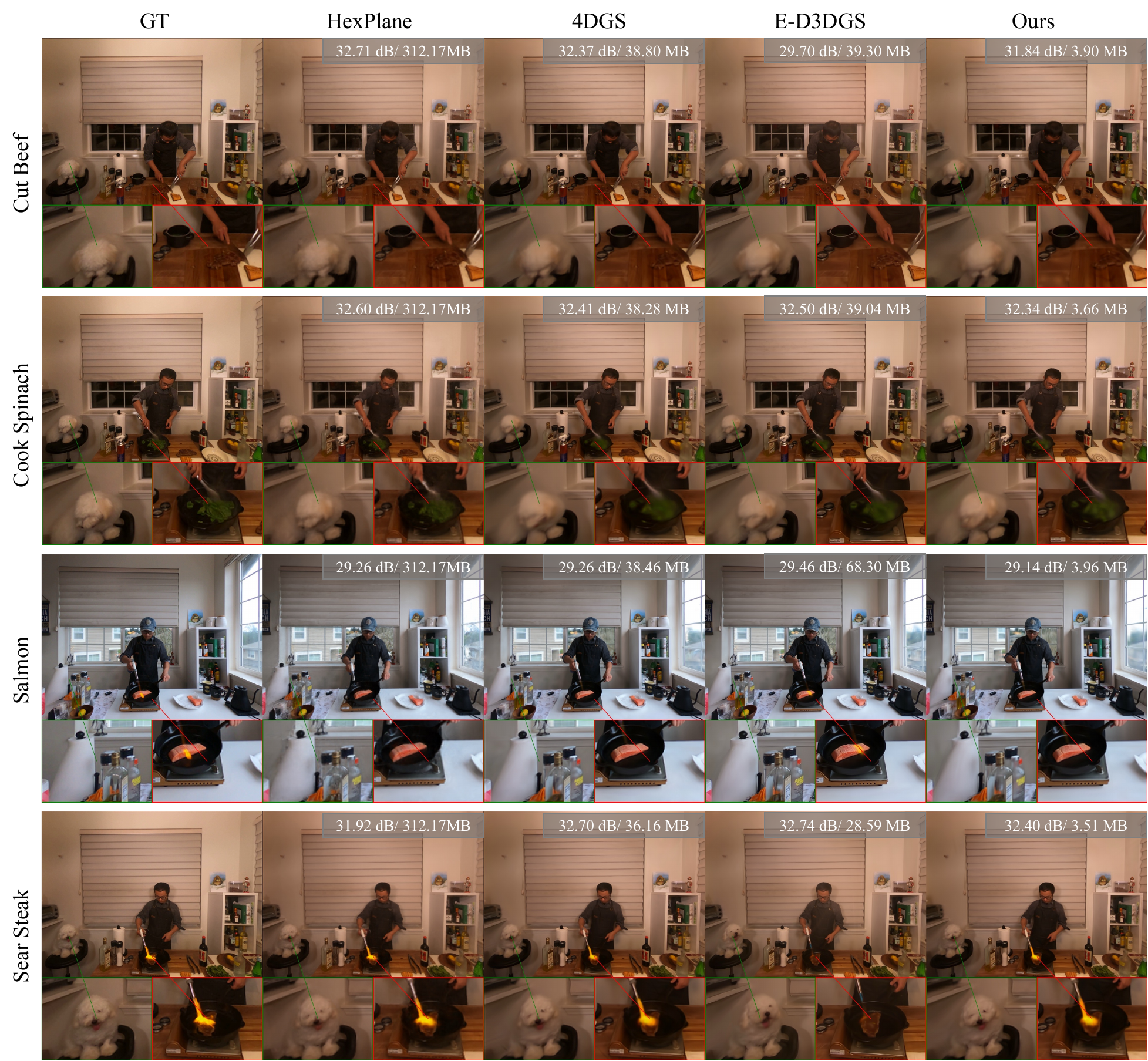}
    \caption{Qualitative quality comparisons of \textit{Standup} in the Neu3D dataset. Light4GS achieves up to \textbf{65-272}× compression with a maximum 1\% PSNR degradation.}
    \label{fig:vis}
    \vspace{-0.5cm}
\end{figure*}

\subsection{Adaptive Quantization}
In deep context modeling, each feature is quantized before being processed by the distribution estimation network. However, conventional deep context modeling uses a fixed quantization step size of 1, which is unsuitable for multiscale hexplanes due to their different value range. Additionally, space-only and space-time planes in 4DGS exhibit distinct means and variances owing to differences in spatial and temporal scales as shown in Fig. \ref{prune}. To address this, we propose an Adaptive Quantization (AQ) method in which different learnable quantization steps \( q_c \) are assigned to the two groups during 4DGS training. According to \cite{balle}, quantization can be approximated as adding uniform noise over the quantization range to enable differentiable training. This is illustrated in the total entropy loss for multiscale hexplanes as:
\begin{equation} R_{\rm Hex} = -\log_2 \left( \prod_{l} \prod_{c} \mathcal{N}(\mathbf{\Phi}_{c}^{l}) * (\mathcal{U}(-\frac{1}{2}, \frac{1}{2})q_c) \right), \end{equation}
where $\mathcal{N}$ calculates the total probability distribution of the feature plane $\mathbf{R}^{l}_c$ based on its predicted parameters, $q_c$ is the learnable quantization step for plane $c$, and $\mathcal{U}$ denotes the uniform distribution with unit range.

\subsection{Entropy-constrained Compression for SH}
Given that SH coefficients account for over 80\% of each primitive's storage, we utilize a fully factorized entropy model \cite{balle} to compress them directly. Specifically, each SH coefficient $c_i$ of a primitive has its probability predicted by the \textit{entropy bottleneck} and is then encoded using AC. The cumulative entropy loss of the SH coefficients $R_{\text{SH}}$ is combined with the hexplane entropy loss $R_{\text{Hex}}$ to form the total entropy loss $L_{\text{entropy}}$.

\noindent\textbf{Training Loss: }The overall training loss comprises three terms: the image reconstruction loss \(L_{\text{PSNR}}\); the total entropy loss \(L_{\text{entropy}}\) and the spatio-temporal regularization term \(L_{\text{ST}}\) from \cite{HexPlane}, which encourages smooth transitions in the deformation field. The total loss is given by:
\begin{equation}
    L = L_{\text{PSNR}} + \lambda L_{\text{entropy}} + \alpha L_{\text{ST}},
\end{equation}
where \(\lambda\) and \(\alpha\) control the trade-off between compression efficiency and temporal smoothness.

\section{Experiment}
In this section, we describe the implementation details, perform comparisons with previous methods on three benchmark datasets, and evaluate the rate-distortion performance of Light4GS along with a comprehensive ablation study. 

\subsection{Experimental Configuration}
Our model is based on the 4DGS framework \cite{4DGS} using PyTorch \cite{pytorch} and is trained on a single NVIDIA RTX 3090 GPU. We have fine-tuned our initialization and optimization parameters based on the configuration outlined in \cite{4DGS}. The implementation of the checkerboard model is based on \cite{compressai} with a 4x downscaling on the hyperprior, and the \textit{entropy model} is adopted from \cite{balle}. During training, we vary $\lambda$ from 5e-5 to 5e-2 and adjust the pruning ratio from 0.19 to 0.64 to achieve different rates. PCA parameters and the hexplane bounds are computed once at the beginning of training from the sparse point cloud. \\
\noindent \textbf{STP implementation: } We implement STP based on \cite{lightGS}, where the significance of each primitive is determined by counting the number of intersected pixels. For non-monocular settings, we randomly select a single camera to estimate primitive significance, as evaluating significance across all cameras is computationally expensive. Notably, STP is applied only after the cloning process for primitives has been completed. To avoid abrupt quality degradation, we perform pruning in two passes: one in the middle of training and another near the end, followed by a few fine-tuning iterations. \\
\noindent \textbf{MHCM Implementation: }We build our deep context model on the \textit{CompressAI} framework \cite{compressai}. The hyperprior encoder $h_a(\cdot)$, decoder $h_s(\cdot)$, and entropy model follow the configuration in \cite{balle}, with a hyperprior that applies $4\times$ downsampling. To extract multiscale deep context, we use a lightweight CNN, where each layer outputs twice as many channels as its input. The checkerboard design with a multiplexer is adopted to accelerate both encoding and decoding \cite{checkerboard}. All hexplane dimensions are chosen as multiples of 4 to ensure compatibility with masked convolutions.

\begin{table}[t]
\centering
\caption{Quantitative results on D-NeRF dataset. The best and the second best results are denoted by \textcolor{red!30}{dark red} and \textcolor{red!15}{light red}.}
\label{dnerf}
\resizebox{\linewidth}{!}{ 
\begin{tabular}{lcccccc}
\toprule[2.5pt]
Model & \textbf{PSNR(dB)$\uparrow$} & \textbf{SSIM$\uparrow$} & \textbf{LPIPS$\downarrow$} & \textbf{Training time$\downarrow$} & \textbf{FPS$\uparrow$} & \textbf{Storage (MB)$\downarrow$} \\
\midrule
TiNeuVox-B \cite{TiNeuVox} & 32.67 & 0.97 & 0.04 & 28 mins   & 1.5  & 48 \\
KPlanes \cite{kplanes}     & 31.61 & 0.97 & -    & 52 mins   & 0.97 & 418 \\
HexPlane-Slim \cite{HexPlane} & 31.04 & 0.97 & 0.04 & 11m 30s & 2.5  & 67 \\
FFDNeRF \cite{ffdnerf}     & 32.68 & 0.97 & 0.04 & -         & $<1$ & 440 \\
MSTH \cite{MSTH}           & 31.34 & \cellcolor{red!30}0.98 & \cellcolor{red!15}0.02 & \cellcolor{red!30}6 mins & -   & - \\
\midrule
3DGS \cite{GS}  & 23.19 & 0.93 & 0.08 & \cellcolor{red!15}10 mins & \cellcolor{red!30}170 & 10 \\
D3DGS \cite{d3dgs} & \cellcolor{red!30}39.11 & \cellcolor{red!30}0.98 & \cellcolor{red!30}0.01 & - & 76 & 37 \\
4DGS-HUST \cite{4DGS}  & \cellcolor{red!15}33.03 & \cellcolor{red!30}0.98 & \cellcolor{red!15}0.02 & 15 mins & 78 & 22 \\
\midrule
\multirow{3}{*}{\textbf{Ours}} 
& 32.58 & 0.96 & 0.03 & 24 mins & \cellcolor{red!15}86 & \cellcolor{red!30}2.20 \\
& 32.79 & 0.97 & 0.03 & 24 mins & 85 & \cellcolor{red!15}2.70 \\
& 33.01 & \cellcolor{red!15}0.98 & \cellcolor{red!15}0.02 & 25 mins & 83 & 3.67 \\
\bottomrule[2.5pt]
\end{tabular}
} 
\vspace{-0.3cm}
\end{table}

\begin{figure*}[t]
   \centering
    \includegraphics[width=0.485\linewidth]{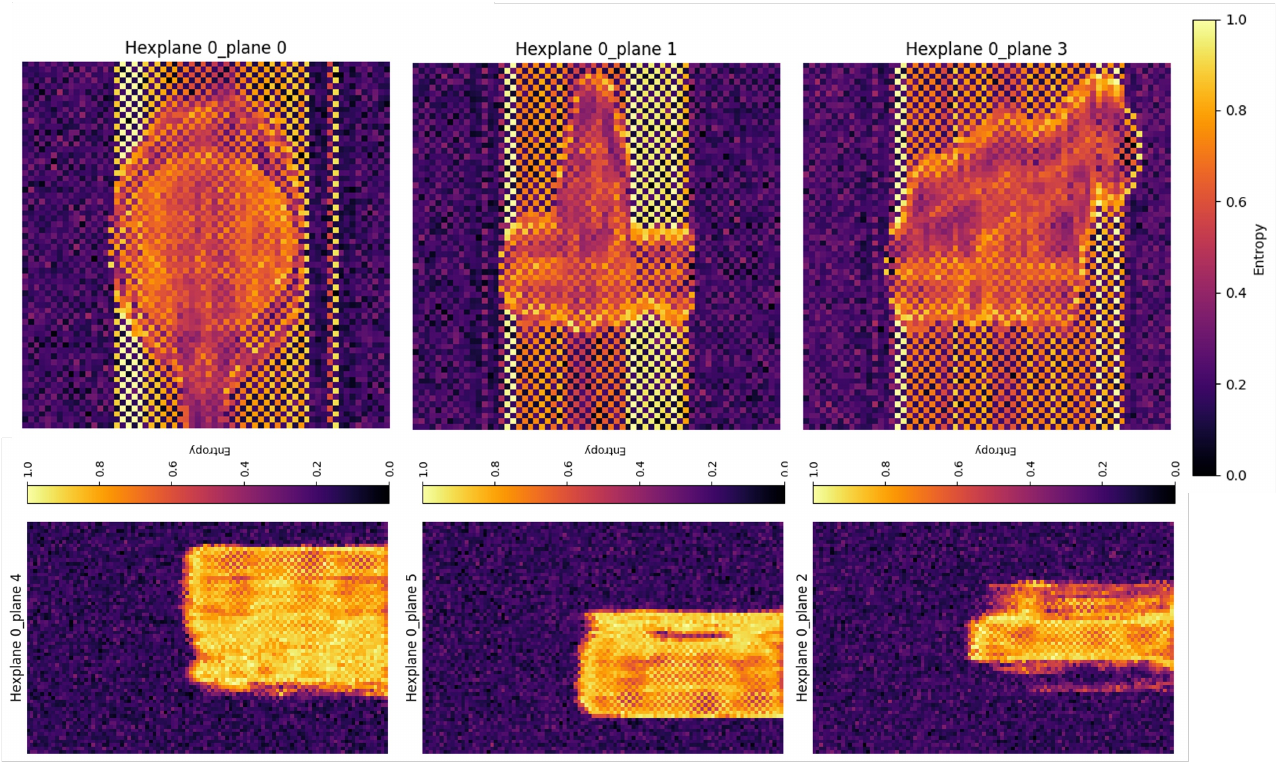}
    \includegraphics[width=0.485\linewidth]{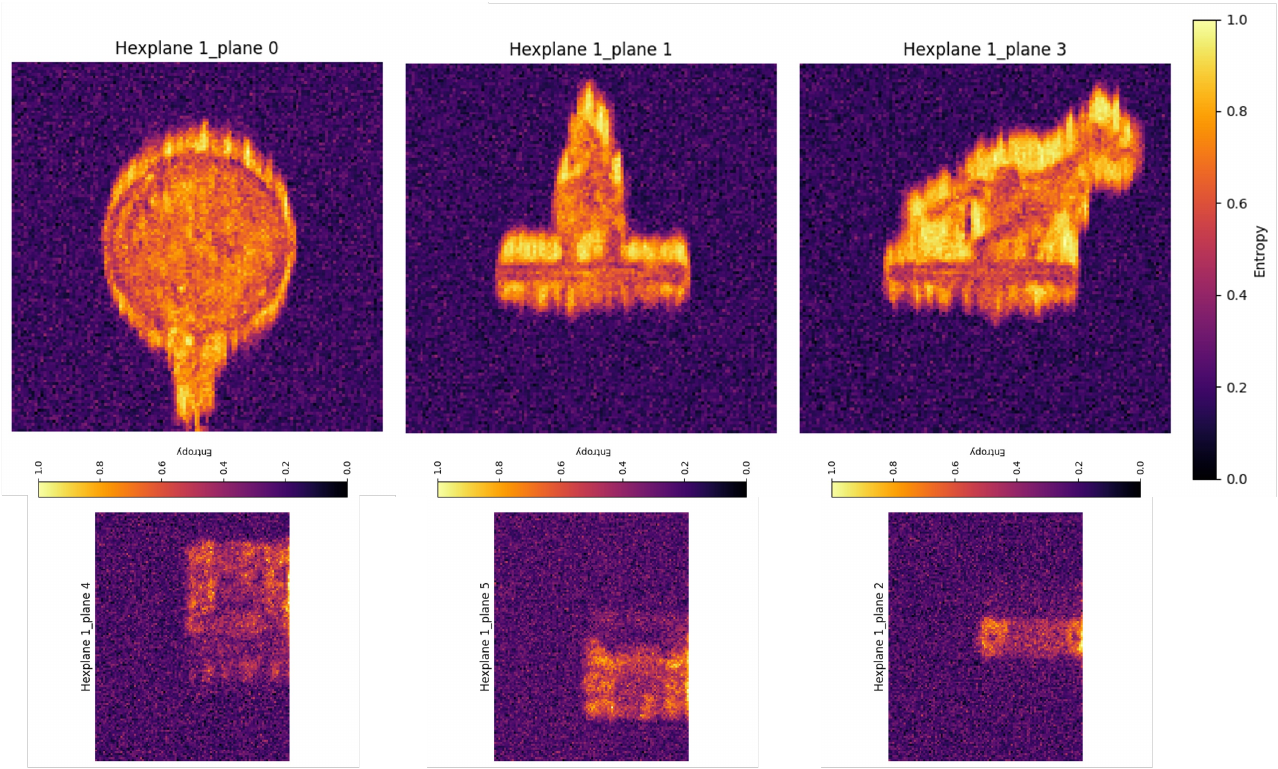}    
    \caption{Bitrate allocation for \textit{Trex} trained at two scales (low scale on the left, high scale on the right). Top row: space-only planes; Bottom row: space-time planes. Bitrate map is normalized.}
    \label{bitratemap}
    \vspace{-0.4cm}
\end{figure*}

\noindent \textbf{Dataset: } For synthetic scene evaluation, we use \textbf{D-NeRF} dataset \cite{dnerf}, which is designed for monocular view synthesis with 50-200 frames from a single camera. For real-world scenes, we use the \textbf{Neu3D}, \cite{neu3d} \textbf{NeRF-DS} \cite{ds} and \textbf{HyperNeRF} \cite{hypernerf} datasets: Neu3D is captured using 15-20 cameras, while HyperNeRF and NeRF-DS employ one or two. Data preprocessing and point cloud initialization are conducted as described in \cite{4DGS}.

\noindent \textbf{Baselines:} We compare our method with recent NeRF-based and GS-based dynamic NVS approaches. \cite{dnerf, ffdnerf, park2021nerfies, hypernerf, TiNeuVox, Mixvoxels, im4d, MSTH, hyperreel} are SOTA NeRF-based approaches. In the GS-based category, methods such as \cite{STG, dynamic3DGS, real-time, 3dgstream} are time-variant, whereas \cite{4DGS, d3dgs, ed3dgs, dn4DGS} are time-invariant representations. Since these approaches lack entropy-constrained optimization, we additionally train the latest entropy-constrained 3DGS method, CompGS \cite{compgs}, across multiple frames to serve as a baseline. During evaluation, some methods are tested on specific datasets, as they are designed exclusively for either monocular or multi-camera setups.

\begin{table}[t]
\centering
\caption{Quantitative results on Neu3D dataset, rendering resolution is set to 1024$\times$768.}
\label{neu3d}
\resizebox{\linewidth}{!}{ 
\begin{tabular}{lcccccc}
\toprule[2.5pt]
Model & \textbf{PSNR(dB)$\uparrow$} & \textbf{D-SSIM$\downarrow$} & \textbf{LPIPS$\downarrow$} & \textbf{Training time} & \textbf{FPS$\uparrow$} & \textbf{Storage (MB)$\downarrow$} \\
\midrule
DyNeRF \cite{dnerf} &  29.58 & 0.020 & 0.083 & - & 0.015 & 28 \\
NeRFPlayer \cite{nerfplayer} & 30.69  & 0.034   & 0.111   & 6 hours   & 0.045   & -  \\
HyperReel  \cite{hyperreel}  & 31.10  & 0.036   & 0.096   & 9 hours   & 2.0     & 360 \\
HexPlane-all$^{\dagger}$ \cite{HexPlane} & 31.70  & \cellcolor{red!30} 0.014   & 0.075   & 12 hours  & 0.2     & 250 \\
Mixvoxels-X \cite{Mixvoxels} &  31.73 & - & 0.064 & - & 4.6 & 500 \\
KPlanes  \cite{kplanes}   & 31.63  & -       & -     & 1.8 hours & 0.3     & 309 \\
Im4D    \cite{im4d}   & \cellcolor{red!30} 32.58  & 0.020   & 0.208   & $\sim$5 hours  & $\sim$5  & 93  \\
MSTH   \cite{MSTH}    & \cellcolor{red!15} 32.37  & 0.015   & 0.056   & \cellcolor{red!30}20 mins  & 2 & 135  \\
\midrule
E-D3DGS \cite{ed3dgs}&  31.20 & 0.026 & \cellcolor{red!30}0.030 & 2h & 42 & 40  \\
STG \cite{STG}& 32.04 & 0.026 & 0.044 & 70 mins & 110 & 175 \\
Dynamic3DGS \cite{dynamic3DGS} & 30.46 & 0.035 & 0.099 & 10 hours & \cellcolor{red!30}460 & 2772 \\
Real-Time4DGS \cite{real-time} & 32.01 & \cellcolor{red!30}0.014 & 0.055 & 9 hours & 114 & $>1000$ \\
3DGStream \cite{3dgstream}& 31.67 & - & - & 60 mins & \cellcolor{red!15}215 & 2340 \\
DN-4DGS \cite{dn4DGS} & 32.02 & \cellcolor{red!30}0.014 & \cellcolor{red!15}0.043 & 50 mins & 15 & 112 \\
CompGS \cite{compgs}& 29.61 & 0.077 & 0.099 & $>1$d & 45 & $\sim825$ \\
4DGS-HUST$^{\dagger}$ \cite{4DGS}    & 31.72  & 0.016   &  0.049   & \cellcolor{red!15} 40 mins   &  34 &  38   \\
SaroGS \cite{yan20244d} & 32.03 & \cellcolor{red!15}0.0142 & 0.044 & $>$2 hours & 40 & 385 \\
4DGS-FDU \cite{ICLR} & 31.57 & 0.016 & 0.057  & $\sim$9 hours & 96 & 3128 \\
4DGC \cite{hu20254dgc} & 31.58 & - & -  & - & 168 & 150 \\
MEGA \cite{zhang2025mega} & 31.49 & 0.016 & 0.056  & - & 77 & 25 \\
Compact3DGS \cite{compactGS} & 31.73 & - & 0.053  & - & 186 & 22 \\
\midrule
\multirow{3}{*}{\textbf{Ours}$^{\dagger}$}      & 31.48  & 0.017   &  0.064   & 1.3 hours   & 40 & \cellcolor{red!30} 3.77   \\
 & 31.62  & 0.017   &  0.057   & 1.3 hours   & 39 & \cellcolor{red!15}4.38   \\
 & 31.69  & 0.016   &  0.053   & 1.3 hours   &  37 &  5.46   \\
\bottomrule[2.5pt]
\end{tabular}
}
\captionsetup{justification=centering}  
\caption*{\tiny Note: $^{\dagger}$ Test excluding \textit{Coffee Martini}. CompGS is trained every 30 frames for estimation.}
\vspace{-0.8cm}
\end{table}

\noindent \textbf{Metrics:} We report the quality of rendered images using peak-signal-to-noise ratio (PSNR), (multiscale) structural similarity index (SSIM), structural desimilarity index (D-SSIM), perceptual quality measure LPIPS, rendering frames per second (FPS), training time and storage. Three rate-distortion points are displayed for each dataset due to space limitation. \\

\subsection{Training Process}
\textbf{D-NeRF \cite{dnerf}:} During the first 3000 iterations, only the standard training process of vanilla 3DGS is applied without any additional techniques. From iterations 3000 to 30000, STP is performed twice based on a predefined ratio, and primitives are densified every 3000 iterations until 14000 iterations. The entropy constraints of the MHCM and SH are applied every 3 iterations, with the SH constraint activated only after densification ends. Adaptive quantization starts at 15000 iterations, with the SH quantization fixed at 0.02 based on pre-evaluation of primitives generated from vanilla 4DGS.\\
\noindent \textbf{Neu3D, HyperNeRF and NeRF-DS \cite{neu3d, hypernerf, ds}: }The training pipeline for the two real-world datasets follows that of D-NeRF, with the only modification being a reduction in training iterations to 20,000.


\begin{table}[t]
\centering
{
\caption{Quantitative results on NeRF-DS's dataset. Rendering resolution is set to 960×540.}
\label{hypernerf}
\resizebox{0.99\linewidth}{!}{ 
\small
\begin{tabular}{lccccc}
\toprule[2.5pt]
Model & \textbf{PSNR(dB)$\uparrow$} & \textbf{SSIM$\uparrow$} & \textbf{FPS$\uparrow$} &  \textbf{Time}  & \textbf{Storage (MB)$\downarrow$} \\
\midrule
3DGS \cite{GS} 
& 20.93 
& 0.7787 
& - 
& 5 mins 
& \cellcolor{red!10}13 \\
D3DGS\cite{d3dgs} 
& \cellcolor{red!10}23.61 
& 0.8395 
& 35 
& - 
& 53 \\
MotionGS \cite{guo2024motion} 
& \cellcolor{red!30}24.54 
& \cellcolor{red!30}0.8656 
& 35 
& - 
& 62 \\
4DGS-HUST \cite{ffdnerf} 
& 23.57 
& \cellcolor{red!10}0.8399 
& \cellcolor{red!10}46 
& 15 mins 
& 24 \\
\midrule
\textbf{Ours}  
& 23.59 
& 0.8367 
& \cellcolor{red!30}61 
& 20 mins 
& \cellcolor{red!30}2.36 \\
\bottomrule[2.5pt]
\end{tabular}
} 
\arrayrulecolor{black} 
}
\vspace{-0.2cm}
\end{table}

\begin{table}[t]
\centering
\caption{Quantitative results on HyperNeRF's dataset. Rendering resolution is set to 960$\times$540.}
\label{hypernerf}
\resizebox{0.99\linewidth}{!}{ 
\small
\begin{tabular}{lcccccc}
\toprule[2.5pt]
Model & \textbf{PSNR(dB)$\uparrow$} & \textbf{MS-SSIM$\uparrow$} & \textbf{Time$\downarrow$} & \textbf{FPS$\uparrow$} & \textbf{Storage (MB)$\downarrow$} \\
\midrule
Nerfies \cite{park2021nerfies} & 22.20 & 0.803 & $\sim$ hours & $<$ 1 & - \\
HyperNeRF \cite{hypernerf} & 22.40 & 0.814 & 32 hours & $<$ 1 & - \\
TiNeuVox-B \cite{TiNeuVox} & 24.30 & 0.836 & \cellcolor{red!30}30 mins & 1 & 48 \\
FFDNeRF \cite{ffdnerf} & 24.20 & 0.842 & - & 0.05 & 440 \\
\midrule
3DGS \cite{GS} & 19.70 & 0.680 & \cellcolor{red!15}40 mins & \cellcolor{red!30}55 & 52 \\
D3DGS \cite{ffdnerf} & 22.40 & 0.612 & - & 22 & 129 \\
E-D3DGS \cite{ed3dgs} & 25.72 & - & 2 hours & 26 & 47 \\
DN-4DGS \cite{dn4DGS} & 25.59 & \cellcolor{red!30}0.861 & 1.2 hours & 20 & 68 \\
4DGS-HUST \cite{4DGS} & 25.58 & \cellcolor{red!15}0.848 & \cellcolor{red!30}30 mins & 22 & 63 \\
4DGS-FDU \cite{ICLR} & \cellcolor{red!15}27.44 & - & - & - & 73 \\
MoDec-GS \cite{kwak2025modec} & \cellcolor{red!30}27.78 & - & - & - & 41 \\
\midrule
\multirow{3}{*}{\textbf{Ours}} & 25.35 & 0.845 & 50 mins & \cellcolor{red!15}28 & \cellcolor{red!30}5.15 \\
& 25.46 & 0.845 & 50 mins & 27 & \cellcolor{red!15}6.84 \\
& 25.55 & 0.847 & 50 mins & 27 & 8.87 \\
\bottomrule[2.5pt]
\end{tabular}
} 
\vspace{-0.2cm}
\end{table}

\begin{figure*}[t]
    \centering
    \begin{minipage}[t]{0.48\linewidth}
        \centering
        \includegraphics[width=0.8\linewidth]{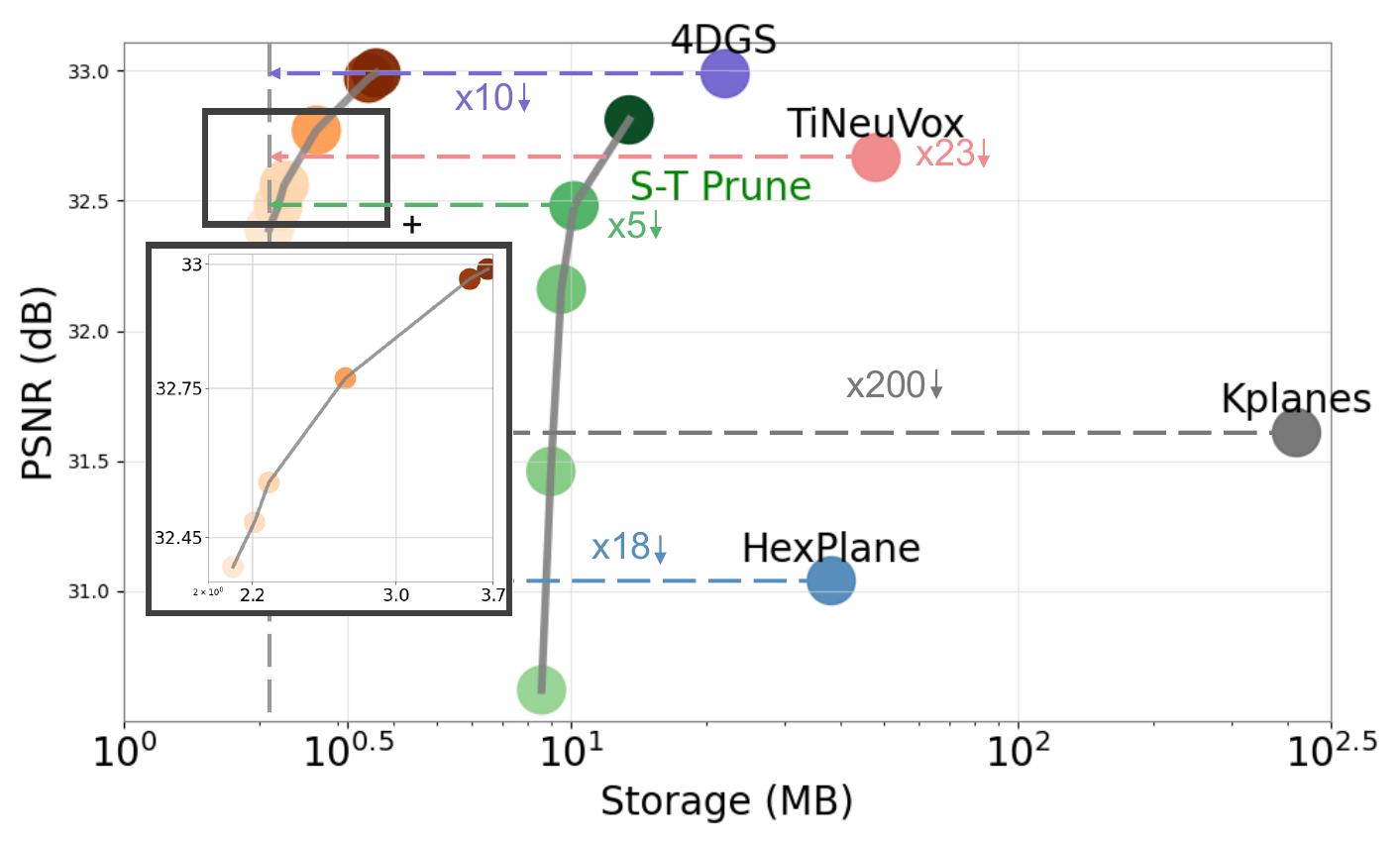}
        \caption{Rate–distortion performance comparison on D-NeRF. Green anchors denote STP only.}
        \label{dnerfrd}
    \end{minipage}
    \hfill
    \begin{minipage}[t]{0.48\linewidth}
        \centering
        \includegraphics[width=0.8\linewidth]{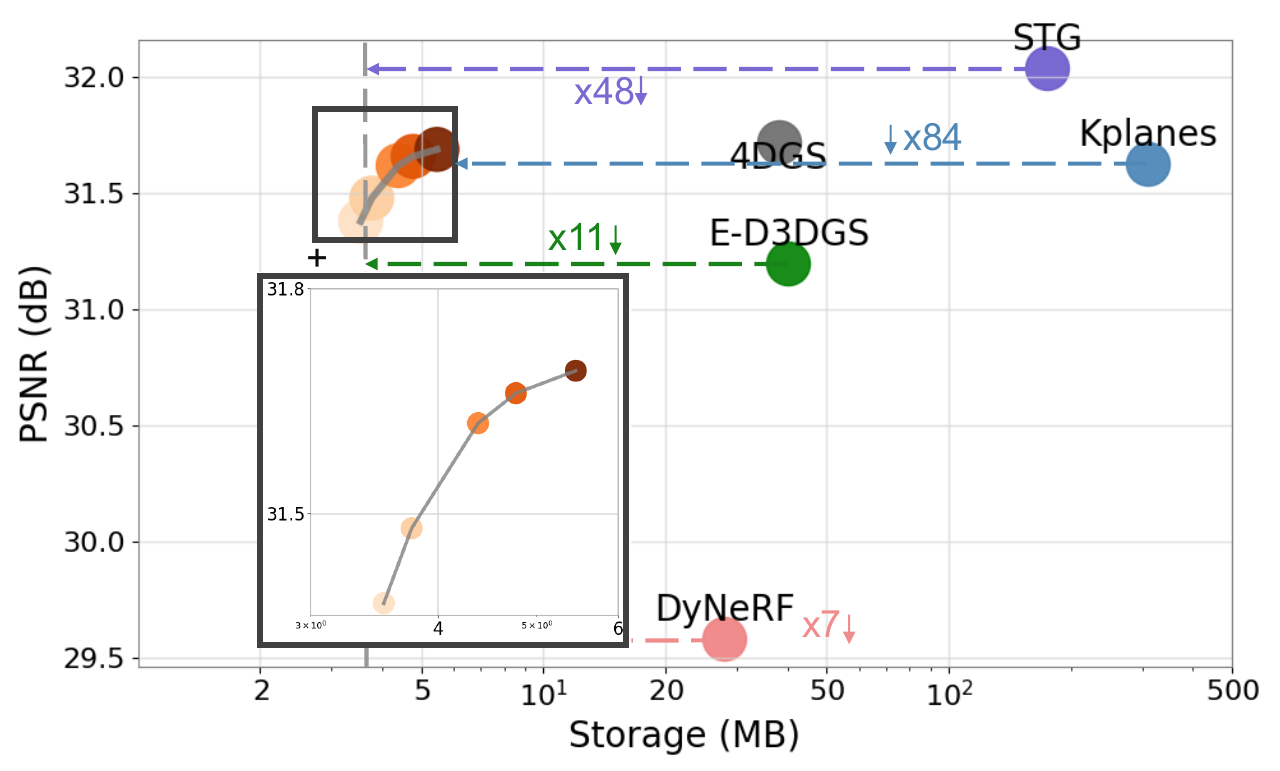}
        \caption{Rate–distortion performance comparison on Neu3D.}
        \label{dynerfrd}
    \end{minipage}

    \vspace{0.3cm}

    \begin{minipage}[t]{0.48\linewidth}
        \centering
        \includegraphics[width=0.8\linewidth]{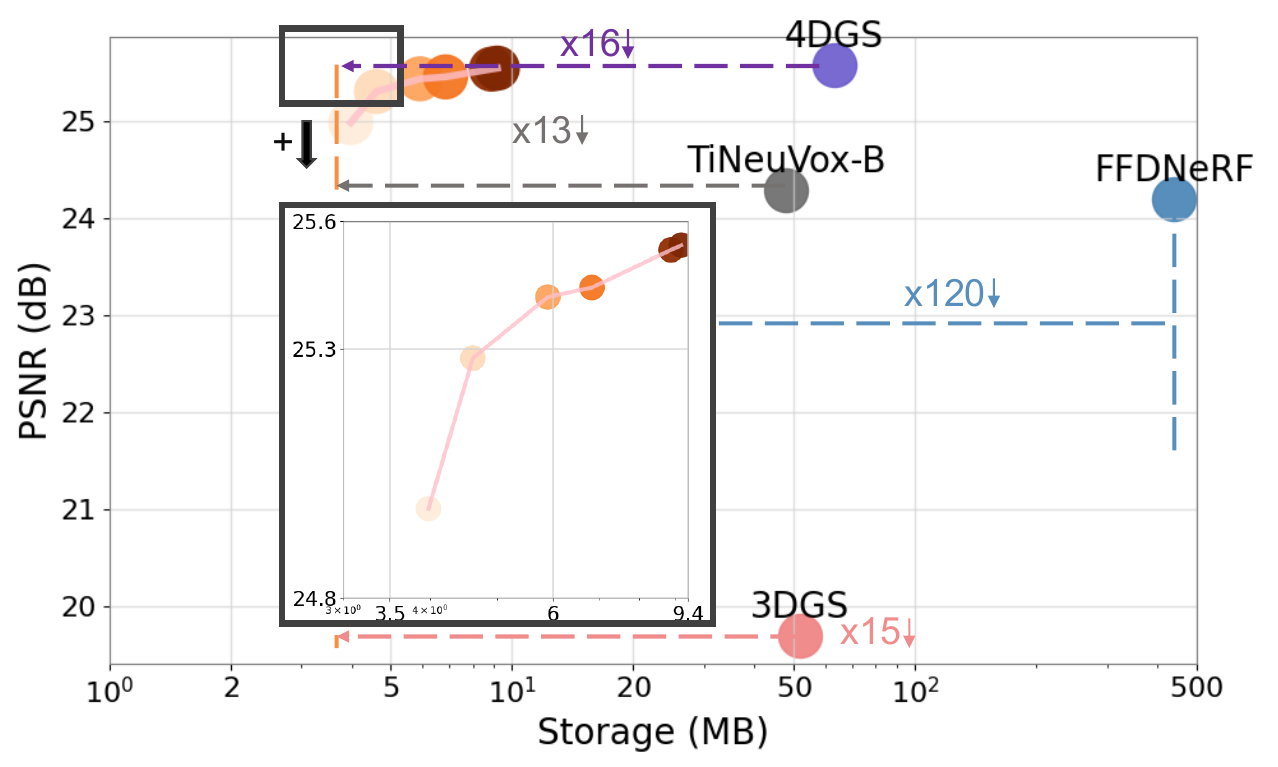}
        \caption{Rate–distortion performance comparison on HyperNeRF.}
        \label{hypernerfrd}
    \end{minipage}
    \hfill
    \begin{minipage}[t]{0.48\linewidth}
        \centering
        \includegraphics[width=0.8\linewidth]{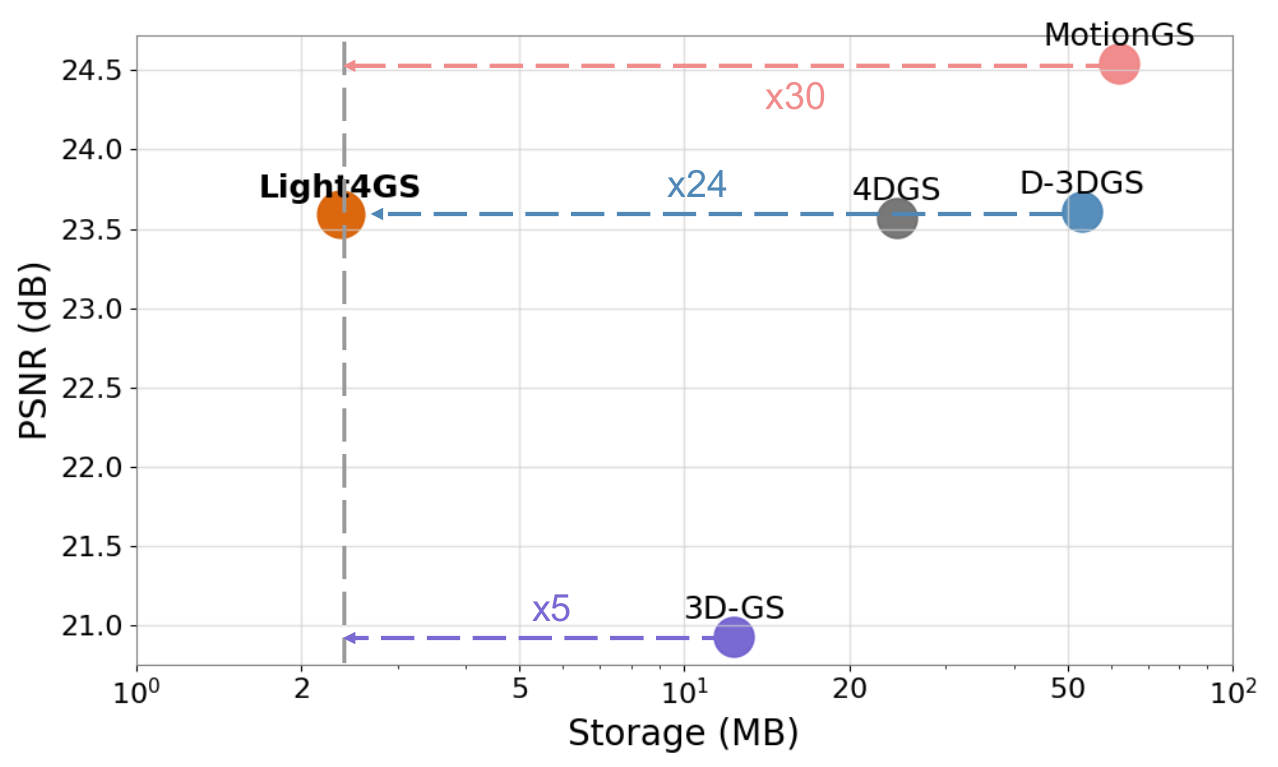}
        \caption{Rate–distortion performance comparison on NeRF-DS.}
        \label{dsnerfrd}
    \end{minipage}
    \vspace{-0.4cm}
\end{figure*}

\subsection{Results}

\textbf{Main Results: }According to the quantitative results in Tables \ref{dnerf}-\ref{hypernerf}, we observe consistent findings across all datasets that: 1) All NeRF-based methods render at extremely low speeds ($<2\text{FPS}$) due to their cumbersome volumetric rendering process. NeRFs using low-rank grid representations \cite{HexPlane, kplanes} perform well in synthetic scenes but struggle with convergence in real-world scenes and require 100–132× more storage than our approach. 2) GS-based methods generally offer improved rendering quality and speed compared to NeRFs;  however, time-variant methods \cite{real-time, 3dgstream, dynamic3DGS} come with significantly larger model sizes and still encounter convergence issues. For example, \cite{dynamic3DGS} demands 735× more storage than our approach, and training \cite{STG, dynamic3DGS, real-time} requires over 8 hours. Deformation-based methods \cite{dn4DGS, ed3dgs, 4DGS} demonstrate better convergence with slightly lower quality, but they still require up to 10× more storage than our approach. 3) Our method outperforms the entropy-constrained GS method, CompGS \cite{compgs}, on Neu3D with a 6.3\% PSNR increase and a 218× storage reduction. This underscores the importance of effective temporal modeling, as CompGS’s intra-entropy constraint restricts its performance compared to most baselines. (4) On the NeRF-DS dataset, our method achieves a strong compression ratio while maintaining comparable quality. Compared with the 4DGS-HUST anchor, our model reduces storage to 2.36~MB, i.e., about $10\times$ compression, with negligible quality change. Since NeRF-DS scenes are longer and exhibit more complex motion, these results further validate the effectiveness of our approach on both short and long real-world sequences. We also provide a qualitative comparison of Light4GS against existing baselines on the Neu3D dataset, where our method achieves the best rate-distortion performance. Please refer to Fig. \ref{fig:vis}. To summarize, Light4GS achieves substantial compression gains not only over NeRF-based and time-variant GS methods but also provides over 10× compression compared to deformable GSs and  200×+ over entropy-constrained GS, without sacrificing other performance metrics.

\begin{figure}[t]
    \centering
    \includegraphics[width=1\linewidth]{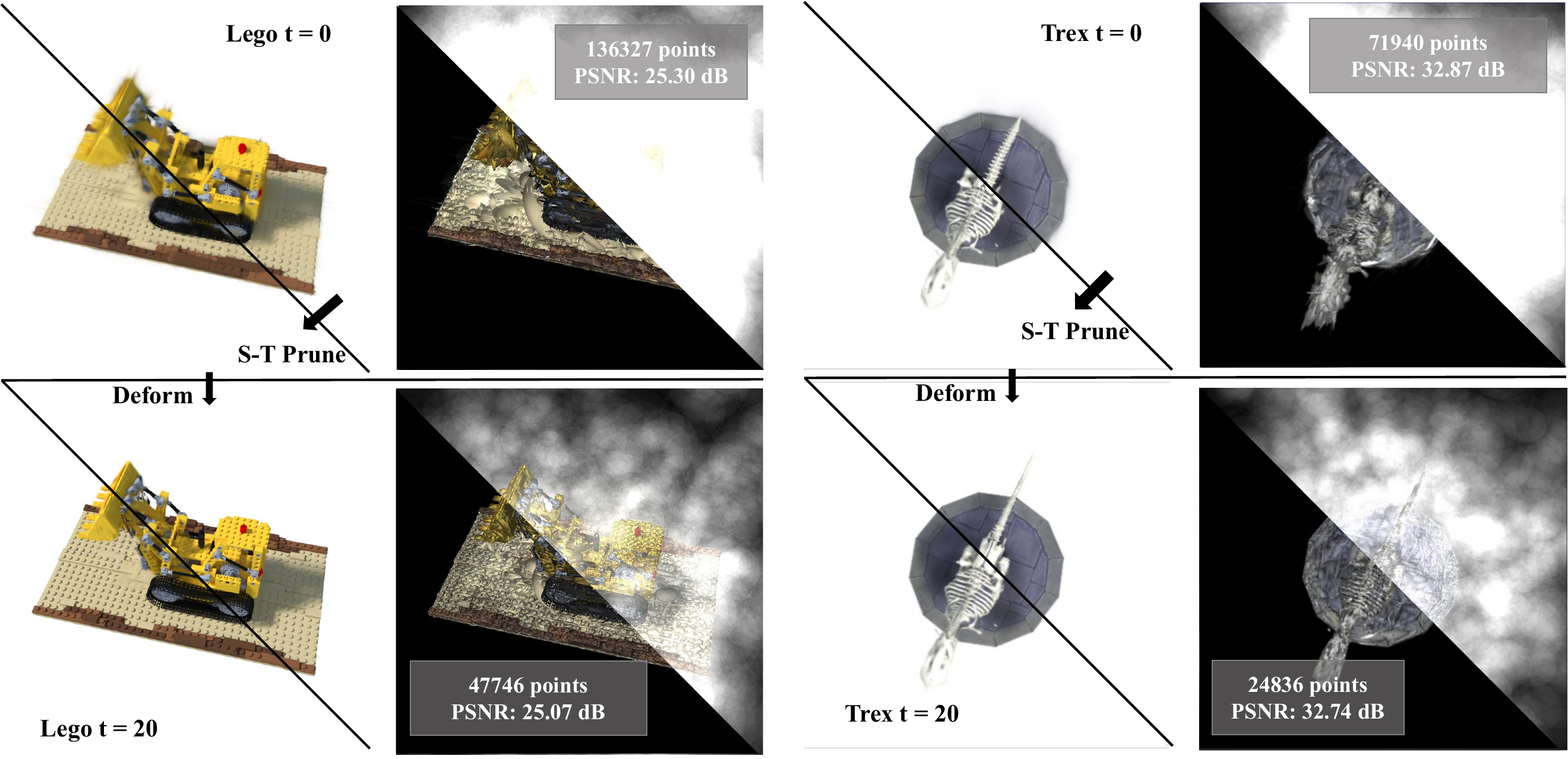}
    \vspace{-0.6cm}
    \caption{Effects of STP, the first (top) and last deformed frame (bottom) is displayed as toy examples.}
    \label{Prune}
    \vspace{-0.3cm}
\end{figure}

\noindent \textbf{Rate Distortion:} The total rate-distortion of Light4GS is determined by three key components: SH coefficients, multiscale hexplanes, and STP. The distortion introduced by STP is controlled by predefined pruning ratios, while the distortion from SH coefficients and multiscale hexplanes is influenced by the parameters $\lambda$. In the experiments, the pruning ratios are set to [0.19, 0.36, 0.51, 0.64], with $\lambda$ selected from [5e-2, 1e-2, 5e-3, 1e-3, 5e-4]. The final rate-distortion curve is constructed by taking the convex hull of all trained rate-distortion points, as illustrated by the solid black line in Fig. \ref{fd}. Fidelity upperbounds, serving as ground truth for rate-distortion performance, are established by training Light4GS at corresponding pruning ratios with $\lambda$ equal to 0.\\
\noindent As shown in Fig. \ref{dnerfrd}-\ref{dsnerfrd}, on D-NeRF, Light4GS achieves over a 10× reduction in storage compared to 4DGS and HexPlane while preserving competitive visual quality. For Neu3D (Fig.~\ref{dynerfrd}), Light4GS demonstrates an 11× reduction in model size relative to 4DGS and achieves an impressive 48× storage reduction over the time-variant 3DGS STG \cite{STG}. For HyperNeRF (Fig.~\ref{hypernerfrd}), Light4GS achieves a 16× storage reduction compared to 4DGS and a 15× reduction relative to per-frame training 3DGS \cite{GS}. In complex scenes, Light4GS presents notable benefits as the requirement for additional scales to capture details increases inter-scale similarities, thus improving compression efficiency.



\begin{figure}[t]
\centering
    \includegraphics[width=0.88\linewidth]{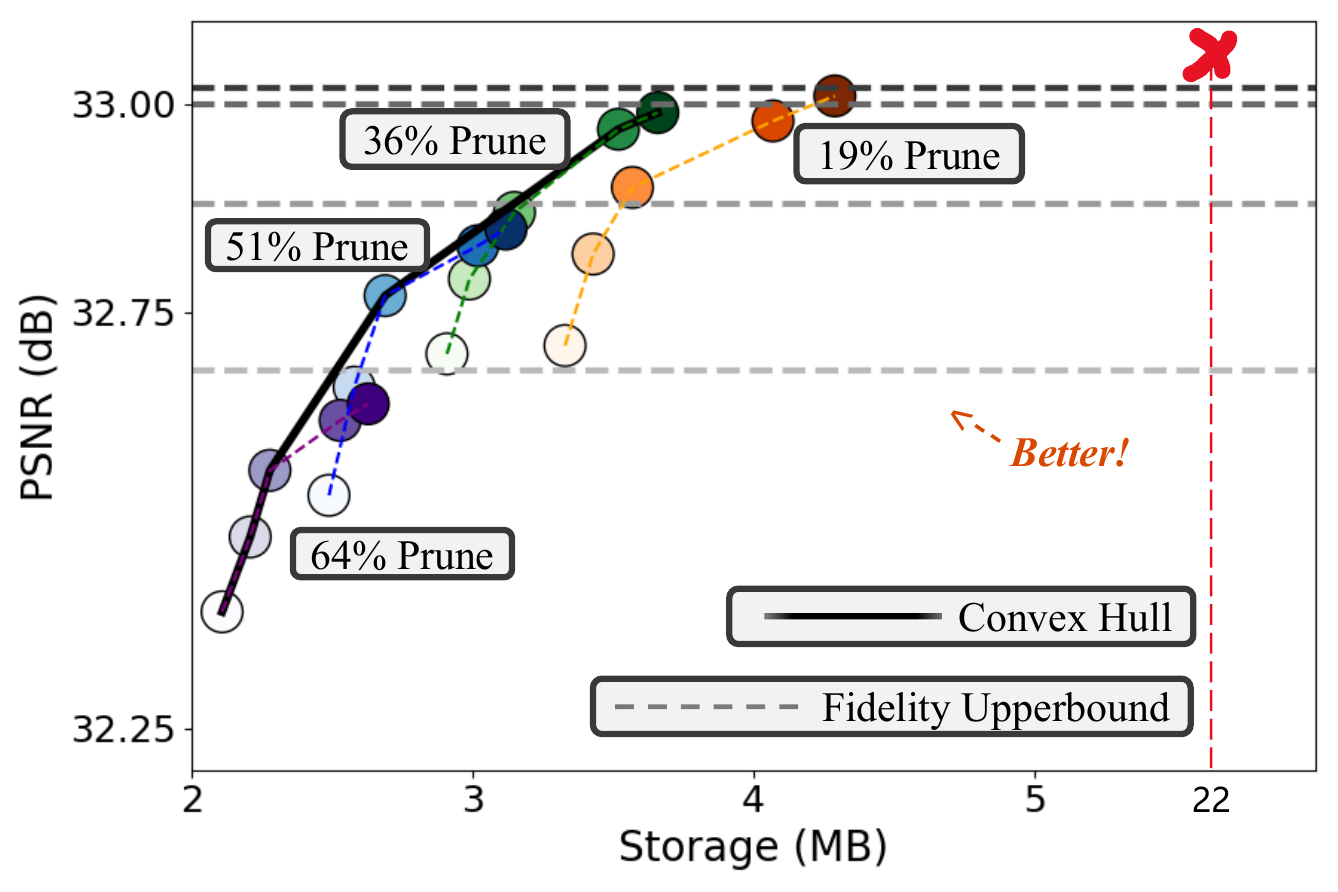}
    \caption{RD-curves and fidelity upperbounds on D-NeRF. STP ratios are denoted by: \textcolor{orange}{Orange} as 19\%, \textcolor{darkgreen}{green} as 36\%, \textcolor{darkblue}{blue} as 51\% and \textcolor{darkpurple}{purple} as 64\%. \textcolor{gray}{Grey} dashed lines represent fidelity upperbounds and \textcolor{red}{Red} cross denotes no STP and rate-distortion.}
    \label{fd}
    \vspace{-0.5cm}
\end{figure}




\noindent \textbf{Bitrate Allocation:} We illustrate how MHCM distributes bits across hexplanes in Fig \ref{bitratemap}. The visualization shows that applying the checkerboard model to the lowest-scale hexplane creates an alternating pattern, where anchor positions efficiently leverage contextual information for optimized bit allocation compared to non-anchor positions. Additionally, the higher-scale hexplane exhibits substantial bitrate reduction, demonstrating that inter-scale context effectively captures cross-scale similarities. As the impact of inter-plane context is minimal and not easily observed, we defer its discussion to the Ablation Study. These observations affirm the effectiveness and rationale behind our MHCM design.

\noindent\textbf{Impact of STP: }In Fig. \ref{Prune}, we present the first and last deformed frames along with their primitive visualizations under a 64\% STP: bottom left is pruned \textit{Trex} vs top right is unpruned \textit{Trex}. Results demonstrate that most primitives have minimal impact on the final rendering. By removing these insignificant primitives, STP produces a more compact representation, i.e., primitives are distributed to form a defined physical shape with respect to spatial content, with only a slight decrease in rendering quality (-0.07 dB). STP also can improve training efficiency, as fewer primitives need optimization and querying to train multiscale hexplanes.



\subsection{Ablation Study}
We conduct detailed introspections of Light4GS's components and rate-distortion primarily on the HyperNeRF dataset. \footnote{If not specified, all ablations are conducted with a 51\% STP ratio.}.


\begin{table}[t]
\centering
{
\caption{Ablation study of \textcolor{red}{MHCM}. "Entropy" denotes entropy-constrained training.}
\label{context abla}
\resizebox{\columnwidth}{!}{
\large 
\begin{tabular}{>{\centering\arraybackslash}p{1.4cm} >{\centering\arraybackslash}p{2.2cm} >{\centering\arraybackslash}p{2.2cm} >{\centering\arraybackslash}p{2.2cm} >{\centering\arraybackslash}p{3.5cm}}
\toprule[2.5pt] 
\textbf{Entropy} & \textbf{Checkerboard} & \textbf{Inter-Scale}& \textbf{Inter-Plane} & \textbf{PSNR} $\uparrow$/\textbf{Size} $\downarrow$ \\ 
\midrule
$\checkmark$ & \textcolor{red}{$\checkmark$} & \textcolor{red}{$\checkmark$} & \textcolor{red}{$\checkmark$} & 0\% / 0\%\\ 
$\checkmark$ & \textcolor{red}{$\checkmark$}& \textcolor{red}{$\checkmark$} & \ding{55} & -0.067\% / +2.31\%  \\ 
$\checkmark$ & \textcolor{red}{$\checkmark$} & \ding{55} & \ding{55} & -0.089\%  / +8.05\% \\ 
$\checkmark$ & \ding{55} & \ding{55} & \ding{55} & -0.215\% / +12.4\% \\ 
\midrule
\ding{55} & \ding{55} & \ding{55} & \ding{55} & +0.426\% / +69.2\% \\
\bottomrule[2.5pt] 
\end{tabular}
}
\arrayrulecolor{black} 
}
\end{table}

\begin{figure}[t]
\centering
    \includegraphics[width=0.85\linewidth]{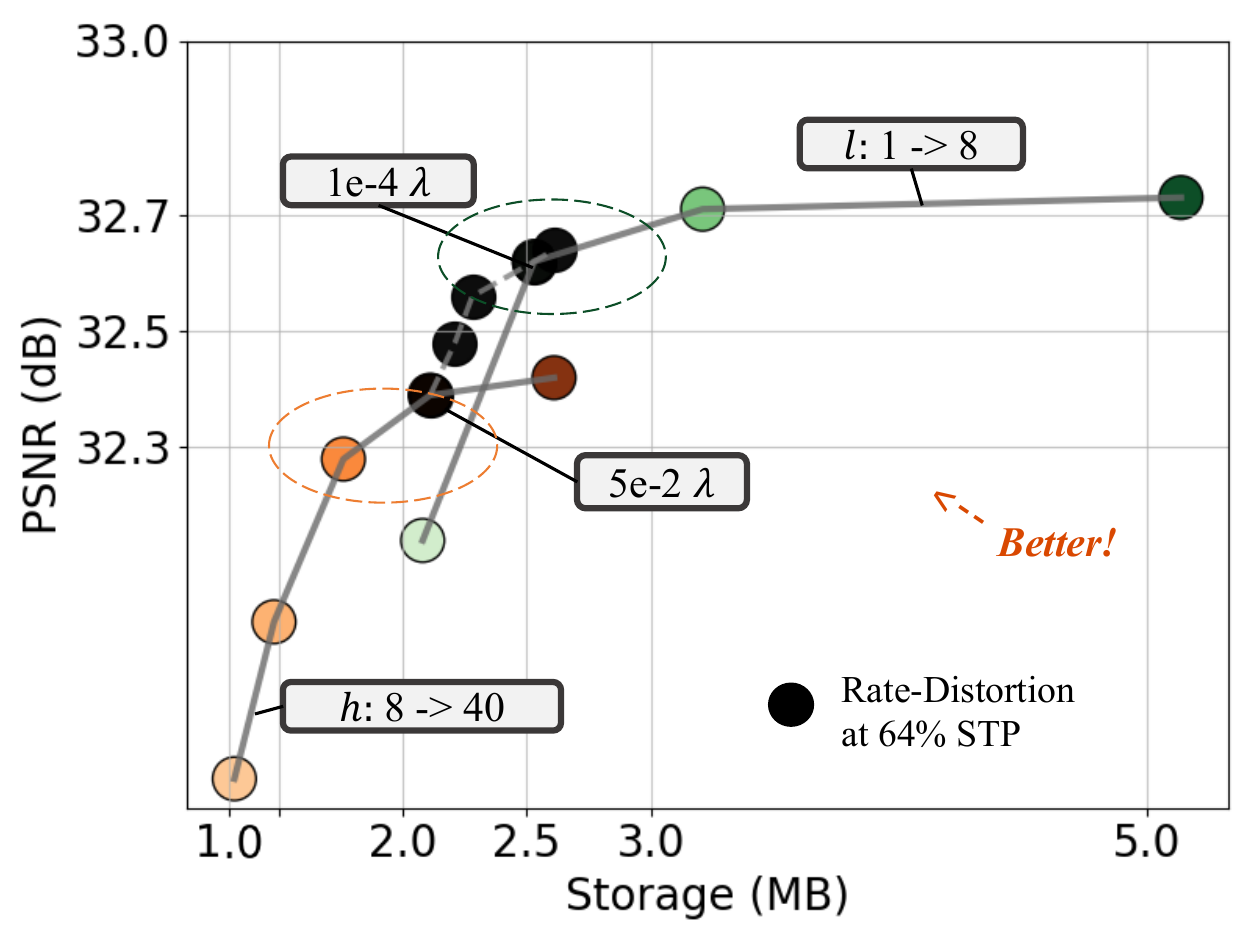}
    \caption{Impact of feature dimension size and multiscale.} 
    \label{feature}
    \vspace{-0.2cm}
\end{figure}

\noindent\textbf{Ablation of MHCM: }As shown in Table \ref{context abla}, integrating all deep contexts of MHCM, i.e., checkerboard, inter-scale, and inter-plane, increases the rate-distortion performance. Inter-plane context has the minimal effect, as it applies only to space-only planes at the lowest-scale. The inter-scale context, however, achieves the largest storage savings by eliminating the storage need for high-scale hyperpriors. Although the checkerboard context provides limited storage savings, because the hexplane rate reduction does not offset the increased network size required for context modeling, it has the greatest impact on quality, yielding a 0.14\% PSNR increase.

\noindent\textbf{Ablation of rate-distortion:} Table~\ref{rd abla} reports an ablation of our MHCM components, where the first row (all enabled) is taken as the reference. Disabling the inter-plane coupling leads to a \(+2.31\%\) size increase with a \(-0.067\%\) PSNR drop, and further removing the inter-scale context enlarges the size to \(+8.05\%\) with a \(-0.089\%\) PSNR drop. When only entropy-constrained training is kept (i.e., without checkerboard / inter-scale / inter-plane), the size increases to \(+12.4\%\) and PSNR decreases by \(-0.215\%\), indicating that the proposed context modules are important for compactness. Finally, removing entropy-constrained training causes a dramatic size increase (\(+69.2\%\)), despite a slight PSNR gain (\(+0.426\%\)), highlighting that entropy-constrained training is the key factor for achieving a favorable rate--distortion trade-off, while checkerboard and cross-scale/plane contexts further improve compression efficiency.

\noindent\textbf{Impact of Hexplane Features: } To explore the impact of scales \( l \) and the hexplane feature size \( h \), we train Light4GS models with different values of \( h \) and \( l \). The black anchors in Fig. \ref{feature} represent the trained rate-distortion curve based on our experimental setting of \( h = 32 \) and \( l = 2 \). For clearer insights, we conduct ablations for the two parameters at different $\lambda$s. Results suggest that our configuration is near the Pareto-optimal region, where increasing the number of scales yields diminishing PSNR gains since higher-scales struggle to capture unique details beyond those at lower-scales, while reducing scales rapidly degrades the rendering quality. A similar trend holds for feature size ablation.

\noindent\textbf{Compressing SH using alternative entropy models:}
Since a purely fully factorized model may leave bits on the table, we also explore three alternative designs for SH compression: (i) a \emph{channel-wise autoregressive} model, where the first three SH channels are encoded with a fully factorized entropy model and an MLP predicts the mean and scale of each subsequent channel conditioned on these three; (ii) a \emph{coupled context} model, where primitive coordinates are used to query hexplane features and the resulting hidden vector is fed into an MLP to predict the conditional Gaussian parameters (mean and scale) shared by all SH channels; and (iii) a \emph{mixture-of-logistics} model, where each channel is modeled as a mixture of several logistic components rather than a single one, using an 8-dimensional parameter vector per channel while keeping all other settings identical to the fully factorized baseline. From the R-D curves on \textit{bouncingballs} (Fig.~\ref{fig:rd}), we observe that only the autoregressive model slightly outperforms the fully factorized baseline at higher bitrates, indicating that explicitly modeling correlations between neighboring SH channels can help at the cost of significant coding delay. The coupled context brings no clear gain, as all channels share a single hexplane feature, limiting its ability to capture channel-wise dependencies and coupling GS decoding with hexplane decoding so that the two can no longer be decoded independently or in parallel. The mixture-of-logistics model, while more expressive in principle, likewise shows no noticeable improvement over the single-logistic fully factorized model in our setting. Considering both compression performance and complexity, we therefore use the fully factorized model for SH in our final implementation.

\begin{figure}[t]
    \centering
    \includegraphics[width=0.8\linewidth]{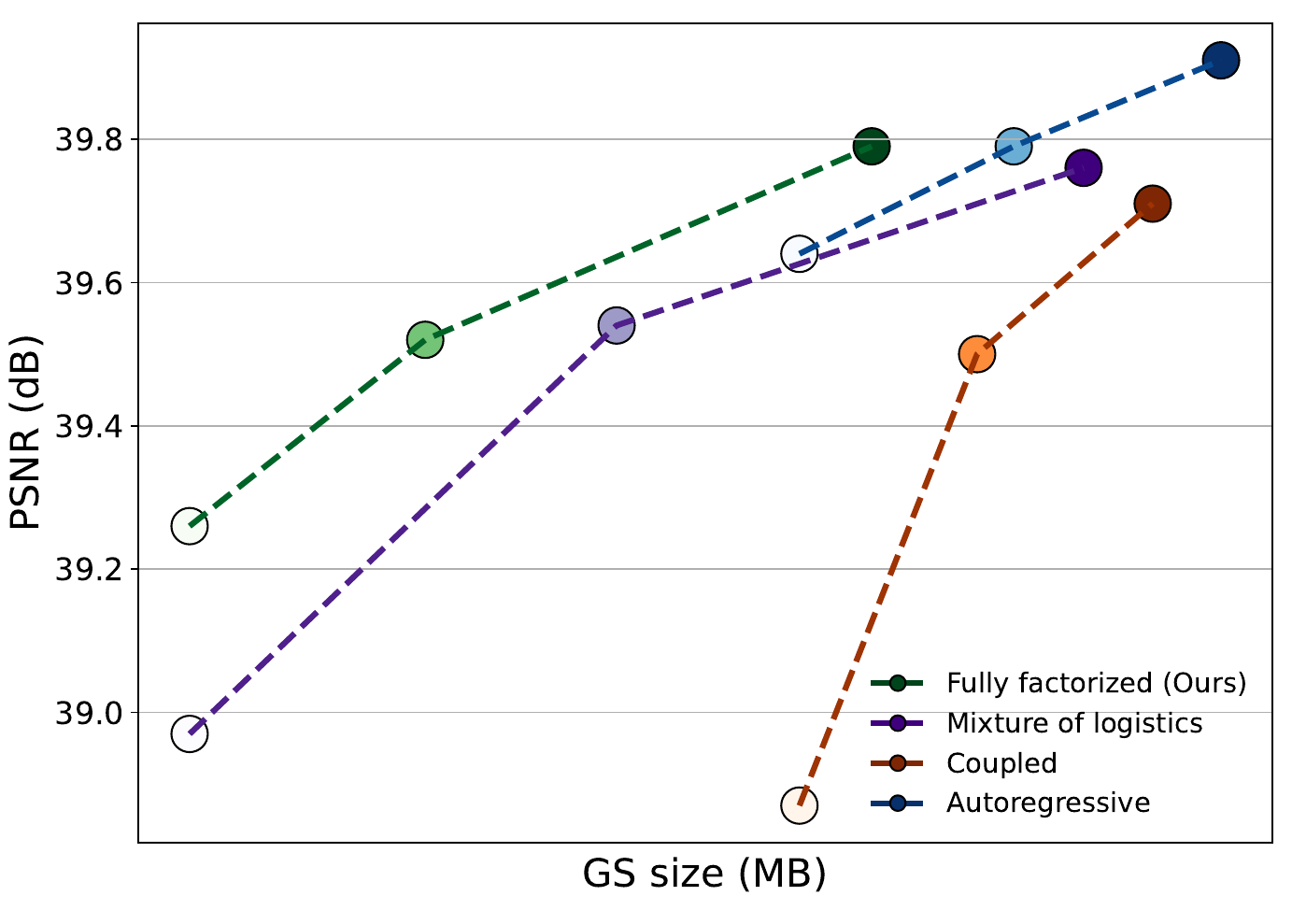}
    \caption{Fully factorized model vs. other entropy models.}
    \label{fig:rd}
\end{figure}

\begin{table}[t]
    \centering
    {
    \caption{Ablation study on PCA, nonlinear contraction, and rate–distortion on HyperNeRF dataset. “GS” denotes Gaussian primitives; “Hex” denotes multiscale hexplanes.}
    \label{rd abla}
    \resizebox{0.9\columnwidth}{!}{
    \large  
    \begin{tabular}{>{\centering\arraybackslash}p{5cm}>{\centering\arraybackslash}p{3.2cm}>{\centering\arraybackslash}p{1.8cm}>{\centering\arraybackslash}p{1.8cm}}
    \toprule[2.5pt]
    \textbf{Ablation items} & \textbf{PSNR}$\uparrow$/\textbf{Storage}$\downarrow$ & \textbf{GS Size}$\downarrow$ & \textbf{Hex Size}$\downarrow$ \\
    \midrule
    Base & 0\%/0\% & 0\% & 0\%\\
    w/o PCA & -1.31\%/- & - & -\\
    w/o No. Contract & 1.23\%/- & - & -\\
    w/ STP & -0.35\%/-15.9\% & -48.3\% & 0\%\\
    w/ STP \& $c_i$-RD  & -0.63\%/-36.5\% & \textbf{-86.5}\% & 0\%\\
    w/ STP \& $\mathbf{R}_i$-RD & -0.70\%/-72.0\% & 62.4\% & \textbf{-95.3}\% \\
    \midrule
    w/ STP \& $\mathbf{R}_i, c_i$-RD & -0.90\%/\textbf{-84.5}\% & \textbf{-88.0}\% &  \textbf{-95.0}\%\\
    \bottomrule[2.5pt]
    \end{tabular}
    }
    \arrayrulecolor{black} 
    }
\end{table}


\begin{figure}[t]
    \centering
    \includegraphics[width=0.99\linewidth]{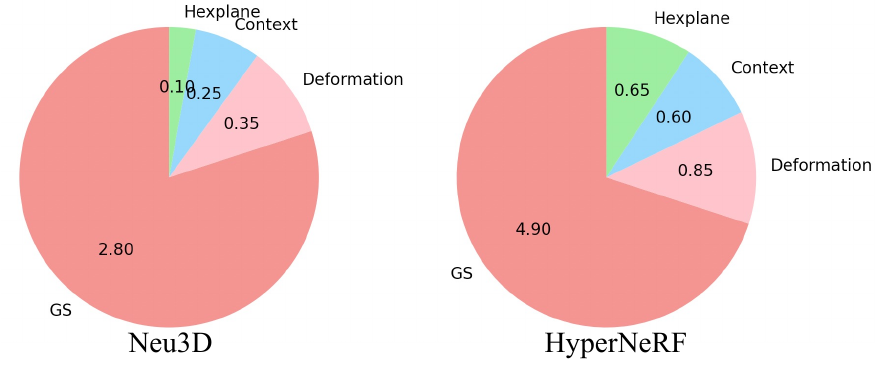}
    \caption{Bitstream composition on Neu3D and DyNeRF (in MB).}
    \label{bitstream}
    \vspace{-0.3cm}
\end{figure}

\begin{figure}[t]
\centering
    \includegraphics[width=0.92\linewidth]{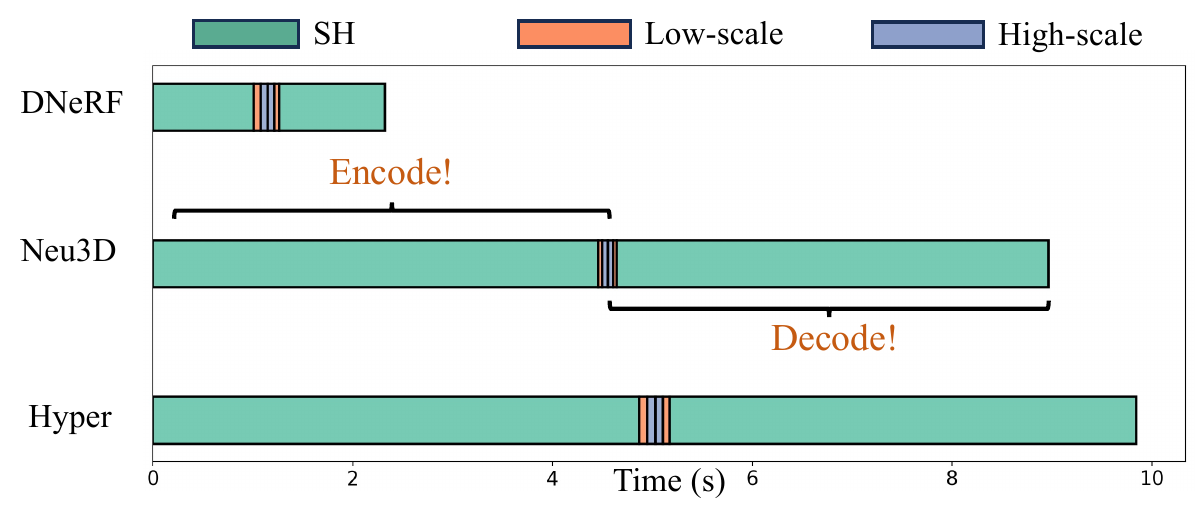}
    \caption{End-to-end coding latency for all scenes. Orange and blue denote coding latencies for hexplanes.} 
    \label{latency}
    \vspace{-0.3cm}
\end{figure}


\subsection{Bitstream Composition}
Light4GS encodes the scene into five main bitstreams: deformable Gaussian primitives (excluding SH), a deformation MLP, entropy-coded multiscale hexplanes, spherical harmonics (SH), and the MHCM module. The MHCM, occupying no more than 0.77 MB in our setup, employs separate CNNs for hyperprior and context extraction, along with distribution prediction. To maintain rendering fidelity, both the MHCM and Gaussian attributes are stored in Float32, with SH compressed via AC based on predicted probabilities. 

Fig. \ref{bitstream} summarizes the bitstream composition across all test scenes, categorized into four components: deformation (MLP), context (MHCM including CNNs and entropy model), hexplane (multiscale features), and GS (Gaussian primitives). Among these, the GS component dominates due to the absence of context modeling in SH compression. The context stream reflects the entire hexplane modeling pipeline, encompassing convolutional, MLP, and entropy coding stages. The relative sizes of deformation and hexplane streams vary with scene complexity, with hexplanes occupying a larger portion in scenes requiring finer spatial detail.

\subsection{Coding Complexity}
We use \textit{torchac} \cite{pytorch} and the Gaussian conditional codec \cite{compressai} to encode the SH coefficients of primitives and multiscale hexplanes. As shown in Fig. \ref{latency}, all scene coding latencies (for either encoding and decoding) remain under 8 ms per frame, enabling real-time streaming for Light4GS. Notably, SH coding operates independently from multiscale hexplane coding, and individual planes of higher-scales can also be coded separately. This independence allows for efficient pipelining of Light4GS coding in practice.

\section{Conclusion}
We propose Light4GS to address the storage and computational demands in dynamic 3DGS, achieving a balanced trade-off between storage efficiency, rendering quality, and computational performance through STP and MHCM. Experiments demonstrate that our approach surpasses existing dynamic NVS methods by delivering higher compression rates with comparable rendering quality. Additionally, entropy-constrained SH compression and adaptive quantization further improve Light4GS's storage efficiency. Since all deformation-based GS models rely on Gaussian primitives and latent embeddings, our method is expected to serve as a benchmark for compressing this type of models in the future.

\section*{Acknowledgment}
This paper is supported in part by National Natural Science Foundation of China (62371290), National Key R\&D Program of China (2024YFB2907204), STCSM under Grant (24511107100), Shanghai Key Lab of Digital Media Processing and Transmission, Shanghai Jiao Tong University.

\bibliographystyle{IEEEtran}
\bibliography{bibtex/bib/IEEEabrv,bibtex/bib/IEEEexample}

\begin{IEEEbiography}
[{\includegraphics[width=1in,height=1.25in,clip,keepaspectratio]{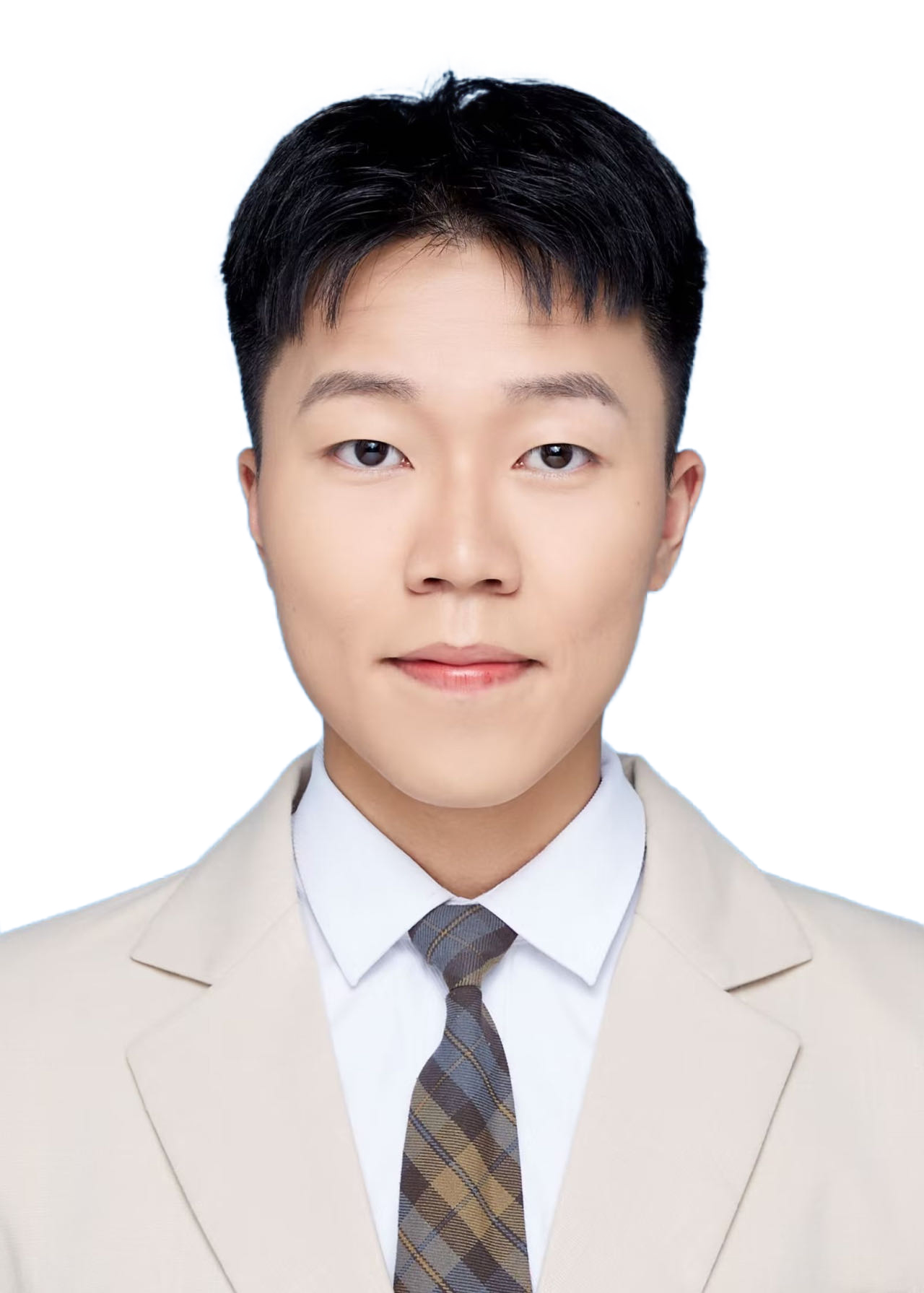}}]{Mufan Liu} received the B.Sc. degree in communication engineering from the University of Electronic Science and Technology of China, Chengdu, China, in 2023. He is currently pursuing the Ph.D. degree with the Cooperative MediaNet Innovation Center, Shanghai Jiao Tong University, Shanghai, China. His main research interests include adaptive streaming, joint source and channel coding and multimedia processing.
\end{IEEEbiography}

\begin{IEEEbiography}
[{\includegraphics[width=1in,height=1.25in,clip,keepaspectratio]{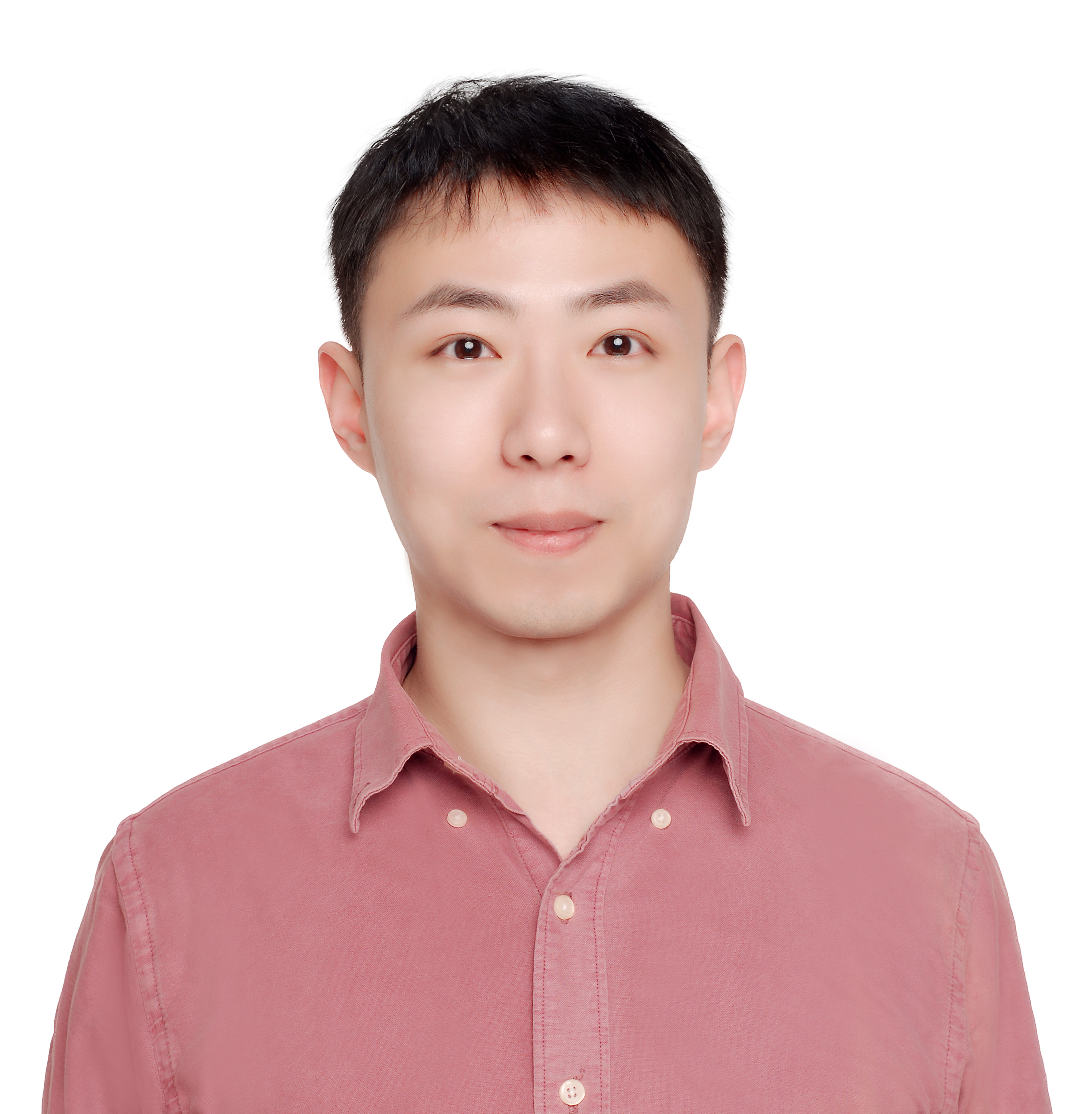}}]{Qi Yang} received the B.S. degree in communication engineering from Xidian University, Xi'an, China, in 2017, and Ph.D degree in information and communication engineering at Shanghai Jiao Tong University, Shanghai, China, 2022. He worked as a researcher in Tencent MediaLab from 2022 to 2024. Now, he joins University of Missouri–Kansas City as a research associate. He has published more than 35 conference and journal articles, including TPAMI, TIP, TVCG, TMM, TCSVT, ICML, CVPR, ICCV, IJCAI, ACM MM, etc. He is also an active member in standard organizations, including MPEG, AOMedia, and AVS. His research interests include 3D/4D GS compact generation and compression, 3D point cloud and mesh quality assessment.
\end{IEEEbiography}

\begin{IEEEbiography}
[{\includegraphics[width=1in,height=1.25in,clip,keepaspectratio]{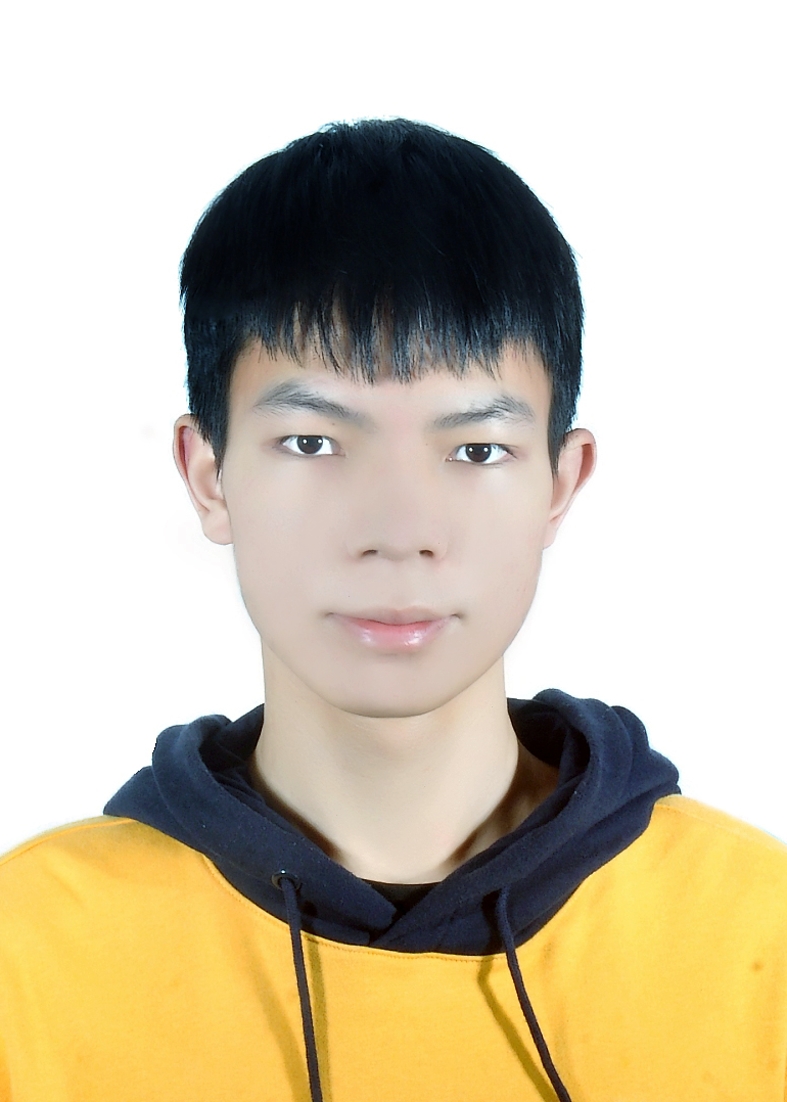}}]{He Huang} received his Bachelor’s degree in communication engineering from Sichuan University, Sichuan, China, in 2023. He is currently pursuing his Ph.D. in Information and Communication Engineering at Shanghai Jiao Tong University, Shanghai, China. His research focuses on 3D immersive media compression, particularly mesh compression and Gaussian Splatting compression.
\end{IEEEbiography}

\begin{IEEEbiography}
[{\includegraphics[width=1in,height=1.25in,clip,keepaspectratio]{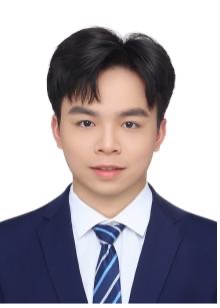}}]{Wenjie Huang} 
Wenjie Huang received the B.Sc. degree in communication engineering from the University of Electronic Science and Technology of China, Chengdu, China, in 2024. He is currently pursuing the Ph.D. degree with the Cooperative MediaNet Innovation Center, Shanghai Jiao Tong University, Shanghai, China. His main research interests include media compression, multimedia processing, channel coding.
\end{IEEEbiography}

\begin{IEEEbiography}[{\includegraphics[width=1in,height=1.3in,clip,keepaspectratio]{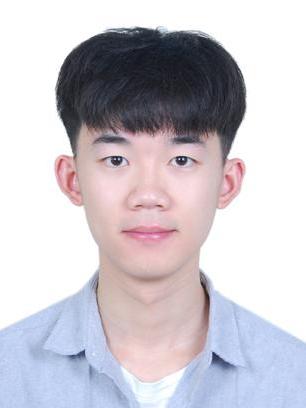}}]{Zhenlong Yuan}
received the B.Sc. degree in telecommunications engineering with management from Beijing University of Posts and Telecommunications, Beijing, China. He is working toward the Ph.D. degree in the Institute of Computing Technology, Chinese Academy of Sciences and University of Chinese Academy of Sciences. His main research interests include vision-language model and 3D reconstruction.
\end{IEEEbiography}

\begin{IEEEbiography}
[{\includegraphics[width=1in,height=1.25in,clip,keepaspectratio]{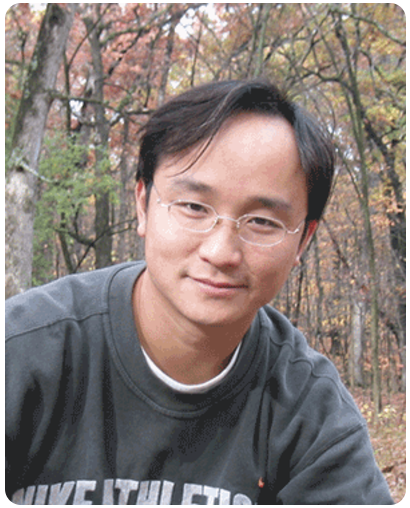}}]{Zhu Li} received the Ph.D. degree in electrical and computer engineering from Northwestern University in 2004. He is a Professor with the Department of Computer Science and Electrical Engineering, University of Missouri, Kansas City, and the Director of NSF I/UCRC Center for Big Learning (CBL) at UMKC. He was an AFRL summer faculty with the UAV Research Center, US Air Force Academy (USAFA), from 2016 to 2018, and from 2020 to 2023. He was Senior Staff Researcher with Samsung Research America’s Multimedia Standards Research Laboratory in Richardson, TX, from 2012 to 2015, Senior Staff Researcher with Future Wei Technology’s Media Laboratory in Bridgewater, NJ, from 2010-2012, Assistant Professor with the Department of Computing, the Hong Kong Polytechnic University from 2008 to 2010, and a Principal Staff Research Engineer with the Multimedia Research Laboratory (MRL), Motorola Labs, from 2000 to 2008. He has over 50 issued or pending patents and over 190 publications in book chapters, journals, and conferences in these areas. His research interests include point cloud and light field compression, graph signal processing and deep learning in the next generation visual compression, image processing and understanding. He is an Associate Editorin-Chief for IEEE TRANSACTIONS ON CIRCUITS AND SYSTEMS FOR VIDEO TECH, Associate Editor for IEEE TRANSACTIONS ON IMAGE PROCESSING (2020), IEEE TRANSACTIONS ON MULTIMEDIA (2015 to 2018), and IEEE TRANSACTIONS ON CIRCUITS AND SYSTEM FOR VIDEO TECHNOLOGY (2016-2019). He received the Best Paper Award at the IEEE International Conference on Multimedia and Expo (ICME), Toronto, in 2006, and the IEEE International Conference on Image Processing (ICIP), San Antonio, in 2007.
\end{IEEEbiography}

\begin{IEEEbiography}
[{\includegraphics[width=1in,height=1.25in,clip,keepaspectratio]{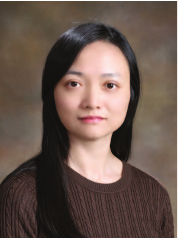}}]{Yiling Xu} received the B.S., M.S., and Ph.D degrees from the University of Electronic Science and Technology of China, in 1999, 2001, and 2004 respectively. From 2004 to 2013, she was a senior engineer with the Multimedia Communication Research Institute, Samsung Electronics Inc., South Korea. She joined Shanghai Jiao Tong University, where she is currently a professor in the areas of multimedia communication, 3D point cloud compression and assessment, system design, and network optimization. She is the associate editor of the IEEE Transactions on Broadcasting. She is also an active member in standard organizations, including MPEG, 3GPP, and AVS.
\end{IEEEbiography}

\begin{IEEEbiography}
[{\includegraphics[width=1in,height=1.25in,clip,keepaspectratio]{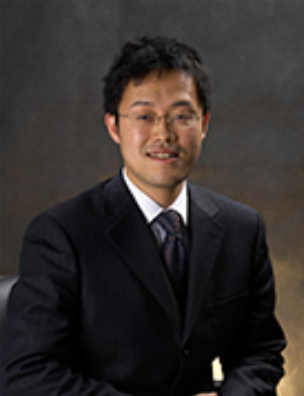}}]{Yunfeng Guan} received the Ph.D. degree from the Department of Electronics and Information Technology, Zhejiang University, Hangzhou, China, in 2003. Since 2003, he has been with the Institute of Wireless Communication Technology, Shanghai Jiao Tong University, where he is currently a Researcher with the Cooperative Medianet Innovation Center. His research interests include HDTV and wireless communications.
\end{IEEEbiography}


\ifCLASSOPTIONcaptionsoff
  \newpage
\fi

\end{document}